\documentclass[10pt,twocolumn,letterpaper]{article}

\usepackage[pagenumbers]{cvpr} %
\usepackage[accsupp]{axessibility}

\definecolor{cvprblue}{rgb}{0.21,0.49,0.74}
\usepackage[pagebackref,breaklinks,colorlinks,allcolors=cvprblue]{hyperref}
\usepackage{tabularx}
\usepackage{array}

\def\paperID{5274} %
\def\confName{CVPR}
\def\confYear{2025}

\title{UrbanCAD: Towards Highly Controllable and Photorealistic 

3D Vehicles for Urban Scene Simulation}

\author{
  Yichong Lu$^{1\text{*}}$
  \and
  Yichi Cai$^{1\text{*}}$
  \and
  Shangzhan Zhang$^{1}$
  \and
  Hongyu Zhou$^{1}$
  \and
  Haoji Hu$^{1}$
  \and
  Huimin Yu$^{1}$
  \and
  Andreas Geiger$^{2,3}$
  \and
  Yiyi Liao\text{$^{1\text{†}}$}\\
  \vspace{-1.5em}
  \and
  $^{1}$Zhejiang University 
  \quad 
  $^{2}$University of Tübingen
  \quad 
  $^{3}$Tübingen AI Center\\
  \href{https://xdimlab.github.io/UrbanCAD/}{\textcolor[rgb]{0,0,1}{\text{https://xdimlab.github.io/UrbanCAD/}}}
}

\newcommand{\bI}{\mathbf{I}}

\newcommand{\bR}{\mathbf{R}}

\newcommand{\bA}{\mathbf{A}}

\newcommand{\bF}{\mathbf{F}}

\newcommand{\bL}{\mathbf{L}}

\newcommand{\cE}{\mathcal{E}}

\renewcommand{\b}{\ensuremath{\mathbf}}

\makeatletter
\DeclareRobustCommand\onedot{\futurelet\@let@token\@onedot}
\def\@onedot{\ifx\@let@token.\else.\null\fi\xspace}

\def\Fig{Fig\onedot}   
\makeatother

\newcommand{\figref}[1]{\Fig~\ref{#1}}
\newcommand{\secref}[1]{Section~\ref{#1}}

\renewcommand{\eqref}[1]{Eq.~\ref{#1}}
\newcommand{\tabref}[1]{Table~\ref{#1}}

\newif\ifcomment
\commenttrue
\ifcomment
	\newcommand{\ag}[1]{ \noindent {\color{red} {\bf Andreas:} {#1}} }
	\newcommand{\yl}[1]{ \noindent {\color{cyan} {\bf YL:} {#1}} }

\else
	\newcommand{\ag}[1]{}
	\newcommand{\yl}[1]{}
\fi

\newcolumntype{P}[1]{>{\centering\arraybackslash}m{#1}}
 
\makeatletter
\def\blfootnote{\gdef\@thefnmark{}\@footnotetext}
\makeatother

\begin{document}
\twocolumn[{%
\renewcommand\twocolumn[1][]{#1}%
\maketitle
\vspace{-3.5em}
\begin{center}
    \centering
    \captionsetup{type=figure}
    \includegraphics[width=0.98\textwidth]{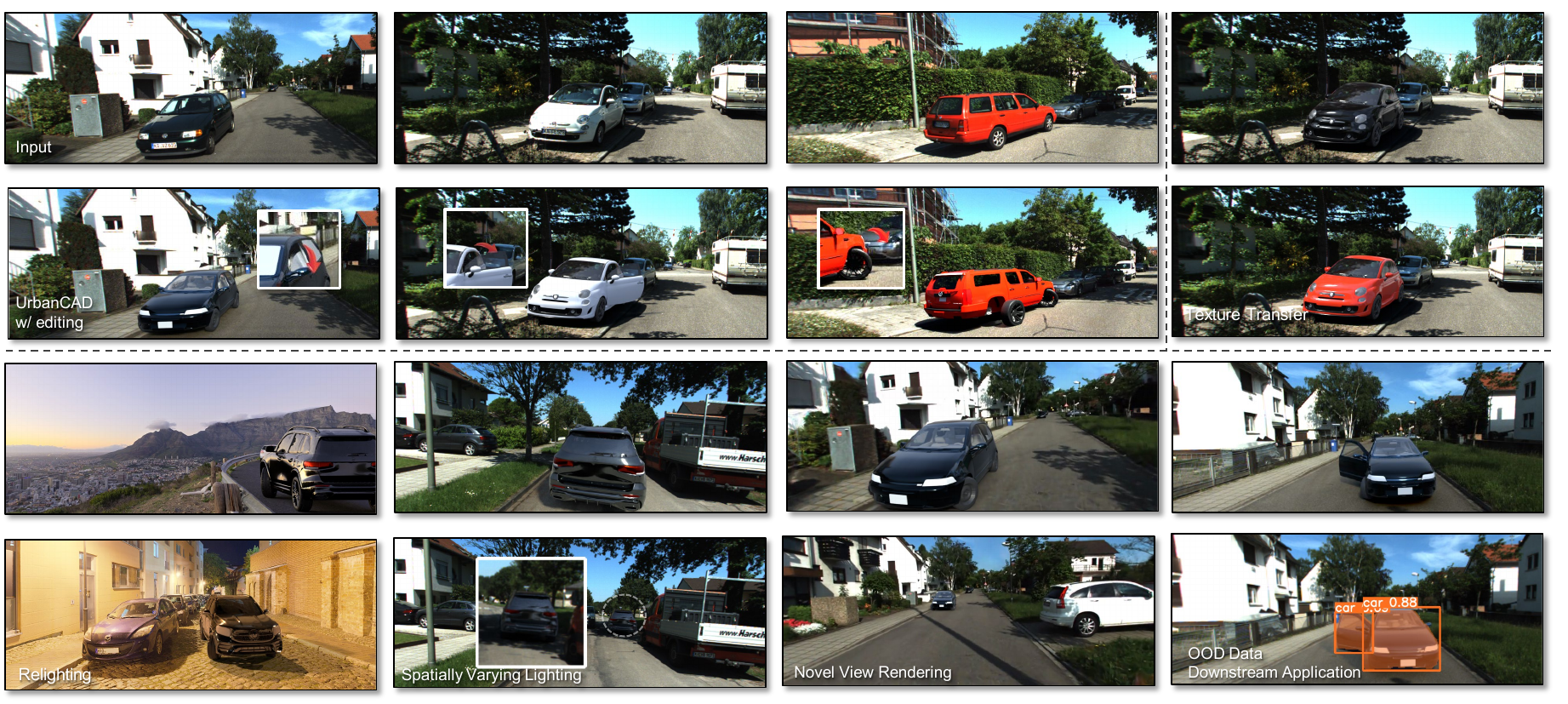}  
    \vspace{-0.2cm}
    \caption{\textbf{UrbanCAD} automatically builds photorealistic and highly controllable digital twins from a single urban image and a large collection of 3D CAD models and handcrafted materials, supporting various editing operations (top). The produced CAD models can be photorealistically inserted into various background scenes and rendered in novel views, synthesizing challenging out-of-distribution (OOD) scenarios with high fidelity for important downstream applications (bottom).}  
    \label{fig:teaser}
\end{center}%
}]

\blfootnote{$^\text{*}$Equal contribution. $^\text{†}$Corresponding author.}
\begin{abstract}
    Photorealistic 3D vehicle models with high controllability are essential for autonomous driving simulation and data augmentation. While handcrafted CAD models provide flexible controllability, free CAD libraries often lack the high-quality materials necessary for photorealistic rendering. Conversely, reconstructed 3D models offer high-fidelity rendering but lack controllability.
    In this work, we introduce UrbanCAD, a framework that generates highly controllable and photorealistic 3D vehicle digital twins from a single urban image, leveraging a large collection of free 3D CAD models and handcrafted materials. 
    To achieve this, we propose a novel pipeline that follows a retrieval-optimization manner, adapting to observational data while preserving fine-grained expert-designed priors for both geometry and material. This enables vehicles' realistic 360$^\circ$ rendering, background insertion, material transfer, relighting, and component manipulation.
    Furthermore, given multi-view background perspective and fisheye images, we approximate environment lighting using fisheye images and reconstruct the background with 3DGS, enabling the photorealistic insertion of optimized CAD models into rendered novel view backgrounds. Experimental results demonstrate that UrbanCAD outperforms baselines in terms of photorealism.
    Additionally, we show that various perception models maintain their accuracy when evaluated on UrbanCAD with in-distribution configurations but degrade when applied to realistic out-of-distribution data generated by our method. This suggests that UrbanCAD is a significant advancement in creating photorealistic, safety-critical driving scenarios for downstream applications.
    \vspace{-2em}
\end{abstract}
    
\section{Introduction} \label{sec:intro}

Photorealistic driving simulators have gained great attention for providing a safe and cost-effective way to evaluate driving algorithms ~\cite{Yang2023UniSimAN,Yang2023ReconstructingOI,Wang2023CADSimRA}. Digital twins of vehicles, representing key traffic participants, are essential for these simulators. Since simulators must assess driving algorithms in both common and rare, long-tailed scenarios, the vehicles within these simulators must exhibit both \textit{photorealism} and \textit{controllability}. 

Classical driving simulators based on game engines, such as CARLA \cite{dosovitskiy2017carla}, use handcrafted CAD models to represent vehicles. While offering high controllability, they suffer from a significant domain gap compared to the real world. Leveraging easily accessible real-world urban images, photorealistic simulation offers a scalable solution to bridge this gap across diverse scenarios.
This direction has gained wide attention with advances in neural rendering techniques~\cite{martin2021nerf,rematas2022urban,tancik2022block,wimbauer2023behind,Sun2024RecentAI}. While these methods close the domain gap and enable photorealistic rendering, they lack fine-grained control over the reconstructed vehicles. \cite{Wang2023CADSimRA} employs CAD models of cars as shape priors and reconstructs vehicles with controllable wheels, appearance, and scene lighting using differentiable rendering. However, it still provides limited control over other vehicle components and yields suboptimal geometry. In addition, vehicles are usually observed from limited viewpoints in urban scenes, impeding the photorealistic rendering of occluded regions. Although prior knowledge can help reconstruct unobserved parts~\cite{Mller2022AutoRFL3,Long2023Wonder3DSI}, the quality of these regions remains unsatisfactory. Furthermore, complex material physical properties such as opacity and metallic reflectance, pose significant challenges for differentiable rendering, especially in our single-view observation setting. In contrast, handcrafted material libraries like Adobe Material Library~\cite{Adobe} provide materials with complex physical properties.
Motivated by these observations, we seek to push the frontier of the photorealism-controllability trade-off.

We move towards this objective by introducing a novel framework that automatically produces 3D vehicles with photorealistic appearance, fine-grained geometric details, and high controllability, including part-level control, from a single urban image and a collection of free 3D CAD models and handcrafted materials. This framework operates within a retrieval-optimization paradigm, performing both CAD retrieval and retrieval-based material optimization. The key idea is to retain the existing detailed expert-designed priors including part-level controllability and complex material physical properties, while refining the appearance to fit the observational data. Specifically, we first retrieve vehicles' geometries represented by handcrafted CAD models, which offer high controllability due to disentangled designs, particularly part-disentangled geometry for component editing. 
Existing reconstruction-based methods~\cite{Wang2023CADSimRA,hong2023lrm,Long2023Wonder3DSI} that use CAD models for training focus only on their appearance or overall geometry, neglecting their disentangled geometry design, which results in a loss of part controllability. Then, we retrieve vehicles' materials represented by optimizable, manually designed procedural graphs, which offer photorealistic material properties—such as opacity and roughness—that are difficult to optimize directly. However, for vehicles that have multiple types of materials, accurately retrieving and assigning the part-aware material priors are not trivial. Previous works~\cite{yeh2022photoscene,shi2020match} typically retrieve material priors based on visual similarity, which is inaccurate for materials with complex physical properties, and primarily focus on objects with a single material. To address this, we retrieve part-aware materials based on semantic meanings via foundation models and assign materials through a ControlNet-based recognition method informed by the retrieved material designs. Finally, we perform part-aware material optimization using physics-based differentiable rendering to align the appearance with the input image.
In addition, given multi-view fisheye and perspective images of the background scene, we propose a fisheye-based spatially varying lighting estimation method to realistically render the optimized CAD model and reconstruct the background using 3D Gaussian Splatting \cite{zhou2024hugs} for high-fidelity novel view background synthesis. The integration of these renderings results in photorealistic novel view synthesis and versatile controllability over foreground vehicles. 

Using this comprehensive pipeline, we systematically evaluate several perception models on our synthesized images. Our experimental results demonstrate that pre-trained perception models retain their performance when replacing real cars with our CAD model renderings for in-distribution data generation. However, they show a clear performance drop when UrbanCAD is used for generating out-of-distribution scenarios, such as cars with opened doors. These results indicate that UrbanCAD produces photorealistic and controllable 3D assets, enabling the creation of rare scenarios for autonomous driving that are not achievable with reconstruction- or retrieval-based methods.

Our main contributions are as follows: 1) We propose a novel pipeline based on the retrieval-optimization paradigm that automatically constructs photorealistic and highly controllable 3D vehicle digital twins with detailed geometry. These digital twins closely align with a single input image and allow for control even over part-level components. %
2) Our system allows for inserting the optimized 3D digital twins back into various urban scenes, and achieving novel view synthesis of the full scene when multi-view images are provided for background reconstruction.
3) We evaluate various vehicle models in terms of fidelity and downstream task accuracy. Our results indicate that our CAD retrieval, material optimization, and lighting estimation modules are all crucial for generating photorealistic out-of-distribution (OOD) scenarios, such as door opening, which are vital for testing the robustness of autonomous perception systems.

\section{Related Work}

\noindent \textbf{Simulation for Autonomous Driving: }
There are two major approaches to sensor simulation for autonomous driving: graphics-based methods~\cite{dosovitskiy2017carla,gaidon2016virtual,shah2018airsim} and data-driven methods~\cite{Wang2021AdvSimGS,Sarva2023Adv3DGS,zhou2024hugs,Ljungbergh2024NeuroNCAPPC,Lindstrom2024AreNR}. Graphics-based simulators, such as CARLA~\cite{dosovitskiy2017carla} and AirSim~\cite{shah2018airsim}, are fast and highly controllable but produce unrealistic simulation results due to substantial manual effort, leading to a significant domain gap for autonomous systems. 
Recently, data-driven methods~\cite{lu2023urban,Guo2023StreetSurfEM,martin2021nerf,rematas2022urban,tancik2022block,wimbauer2023behind,zhang2023nerflets,Liu2023RealTimeNR, Kundu2022PanopticNF, Yang2023UniSimAN,Pun2023LightSimNL, Wang2023NeuralFM, Lin2023UrbanIRLU} have made significant progress in realistic novel view synthesis using neural fields. However, most of these methods have limited editing capabilities and yield suboptimal results when viewing from a large range of angles due to limited observation data. 
Some approaches like~\cite{Wang2023CADSimRA} represent vehicles with mesh and model the wheels separately, allowing for wheel rotation during simulation. However, optimizing the geometry, material, and lighting together in an end-to-end manner is challenging and this design still lacks full controllability over other vehicle components, e.g., windows and doors. In contrast, our CAD model retrieval and optimization-based approach yields a good trade-off between photorealism and controllability.

\noindent \textbf{CAD Model as Scene Representations: } CAD model retrieval has been investigated in many existing approaches~\cite{Avetisyan2018Scan2CADLC, Gao2023DiffCADWP, Gmeli2021ROCARC, Kuo2021Patch2CADPE, Tatarchenko2019WhatDS,gaidon2016virtual}. 
While obtaining good geometry details, the appearance of retrieved CAD models is often unsatisfactory because of the lack of optimization. 
Another line of works~\cite{Engelmann2017SAMPSA,Uy2020DeformationAware3M,Wang2023CADSimRA} utilizes the CAD models as priors and performs geometry optimization afterward. While the optimized geometry is closer to the observation, the CAD models are converted to other scene representations, e.g., implicit surfaces, to allow for optimization, hence losing controllability over vehicle components. In contrast, we retain the detailed geometry and high controllability of CAD models while achieving photorealistic appearance. Concurrently, ACDC~\cite{Dai2024ACDCAC} obtains the digital cousins via CAD retrieval, but it doesn't perform material optimization and lighting estimation for photorealistic rendering to further reduce the domain gap.

\noindent \textbf{Material Transfer from Images: } Recently, image-based mesh texturing methods~\cite{Metzer2022LatentNeRFFS,Bokhovkin2023Mesh2TexGM,Yeh2024TextureDreamerIT,Zeng2023Paint3DPA} using generative models have demonstrated strong performance. However, these methods typically rely on per-vertex texture maps for material representation, which can lead to slow optimization processes. Conversely, we employ procedural graphs to represent materials, resulting in higher quality and faster optimization speeds. Besides, these methods often suffer from multi-face or blurry problems whereas our approach achieves fine-grained and photorealistic materials through effective retrieval and optimization techniques. Another line of work~\cite{Yan2023PSDRRoomSP,yeh2022photoscene} using optimizable procedural graphs mainly focuses on objects with a single material, such as furniture. In contrast, our method extends its capabilities to objects with complex materials, such as vehicles, by part-aware material retrieval.

\section{Vehicle CAD Retrieval and Optimization} \label{sec:method1}

Our method begins with CAD retrieval and optimization, using a single urban image and a large collection of free CAD models and material graphs as input. Our aim is to create digital twins that match the reference vehicles in the real-world images, both in geometry and appearance. While these free CAD models are handcrafted with animatable components and detailed geometry, they often lack the high-quality materials required for photorealistic rendering.

\begin{figure*}[!t]  
\centering 
\includegraphics[width=1.\linewidth]{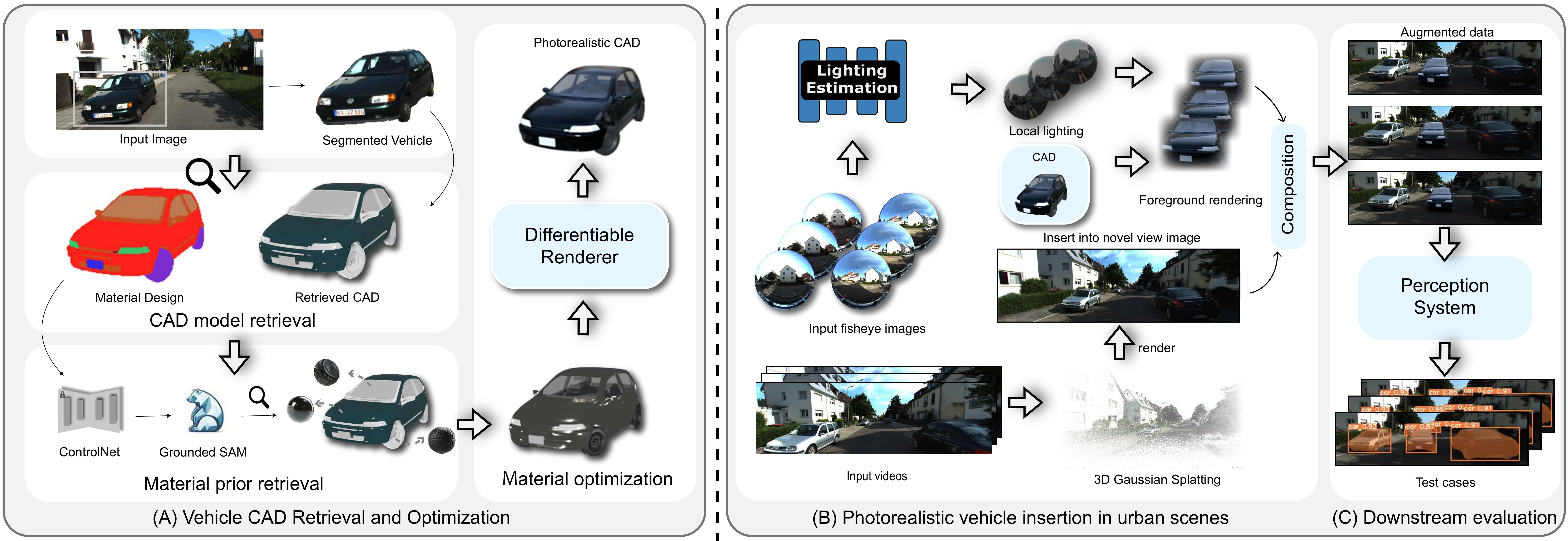}
\caption{ \textbf{Overview of UrbanCAD.} Given a single view input image, we first perform CAD model retrieval and retrieval-based material optimization to create photorealistic and highly controllable vehicle digital twins (left). Given multi-view background images, we then perform realistic vehicle insertion to create various synthetic data for self-driving system testing (right). } 
\label{fig:pipline}
\vspace{-1em}
\end{figure*}

The process consists of three stages, as illustrated in \figref{fig:pipline}. First, we perform image-based CAD model retrieval given a single view input image (\secref{sec:retrieval}). Next, we perform part-aware material prior retrieval using vision foundation models (\secref{sec:mat_retrieval}) and refine the material quality through part-aware optimization (\secref{sec:optimization}).

\subsection{CAD Model Retrieval} \label{sec:retrieval}

Given a single urban image ${\bI}_{input}$ and one user-selected 2D point inside a target vehicle, we segment the reference vehicle image $\bI_{ref}$ from the input scene using SAM~\cite{Kirillov2023SegmentA}. Note that vehicle segmentation can also be automatically obtained via foundation models like GroundedSAM~\cite{Ren2024GroundedSA}. Next, each segmented vehicle image $\bI_{ref}$ is encoded into latent code $\bL_{ref}$ using a pre-trained image encoder $\cE_{clip}$~\cite{Ramesh2022HierarchicalTI}: $\bL_{ref} = \cE_{clip} ( \bI_{ref} )$. Then, we compute latent codes $\left\{\bL_{cad}^i\right\}_{i=1}^N$ for all CAD models \(\{M_{\text{cad}}^i\}_{i=1}^N\) in our library using a pre-trained multi-modality aligned 3D encoder $\cE_{3d}$~\cite{liu2024openshape} : $\left\{\bL_{cad}^i\right\}_{i=1}^N=\cE_{3d} ( \{M_{\text{cad}}^i\}_{i=1}^N )$. Here, $\cE_{3d}$ maps both images and 3D shapes to a shared latent space, where a closer distance in this latent space indicates greater semantic similarity. These latent codes can be pre-cached for efficiency. Finally, we compare the latent codes of the input vehicle images with those of the CAD models using cosine similarity. We identify the CAD model with the highest cosine similarity to the input vehicle image by solving: \(\underset{i}{\operatorname{arg\,max}}\,(\text{sim}(\bL_{\text{ref}}, \{\bL_{\text{cad}}^i\}_{i=1}^N))\). Note that this kind of retrieval can obtain CAD models with the highest semantic similarities aligned with the input image while preserving the handcrafted priors including flexible controllability, detailed geometry, and symmetric material design.

\subsection{Material Prior Retrieval} \label{sec:mat_retrieval}

Adobe Material Library~\cite{Adobe} offers a rich collection of high-quality handcrafted procedural material graphs, which can serve as effective material priors. Consequently, we begin with material prior retrieval before proceeding to material optimization. Previous works ~\cite{yeh2022photoscene,Yan2023PSDRRoomSP} focused on objects with single materials and retrieved material categories based on visual similarity. However, this type of retrieval can be inaccurate, particularly for objects such as vehicles, which are composed of various materials, including glass, where color representation can be ambiguous. To address this problem, we propose retrieving part-aware material priors based on the semantic characteristics of CAD model parts, e.g., windows, wheels, and car bodies.

\noindent \textbf{Material Prior: Optimizable Procedural Node Graphs.} Procedural node graphs $\b G$ provide an expressive material representation in graphics. Unlike per-pixel material parameter maps, these graphs can compactly represent various materials using a small amount of parameters. MATch~\cite{shi2020match} proposes converting such node graphs into differentiable programs, utilizing differentiable rendering to optimize continuous node parameters in an end-to-end manner through rendering loss. The discrete parameters and graph structure, designed by artists, remain fixed.
In this work, we first collect handcrafted procedural material graphs from Adobe Material Library and rename them to the corresponding CAD model part names. For instance, when retrieving material priors for vehicles, we collect three artist-designed base node graphs — glass, rubber, and reflective metal — and rename them to windows, wheels, and car bodies. Note that this process typically needs to be conducted only once for most objects within a given category, as they often share a common set of materials. Then, we follow MATch to translate the handcrafted graphs into optimizable ones to fit the observation data efficiently.

\begin{figure}[t]
    \centering
    \includegraphics[width=0.98\linewidth]{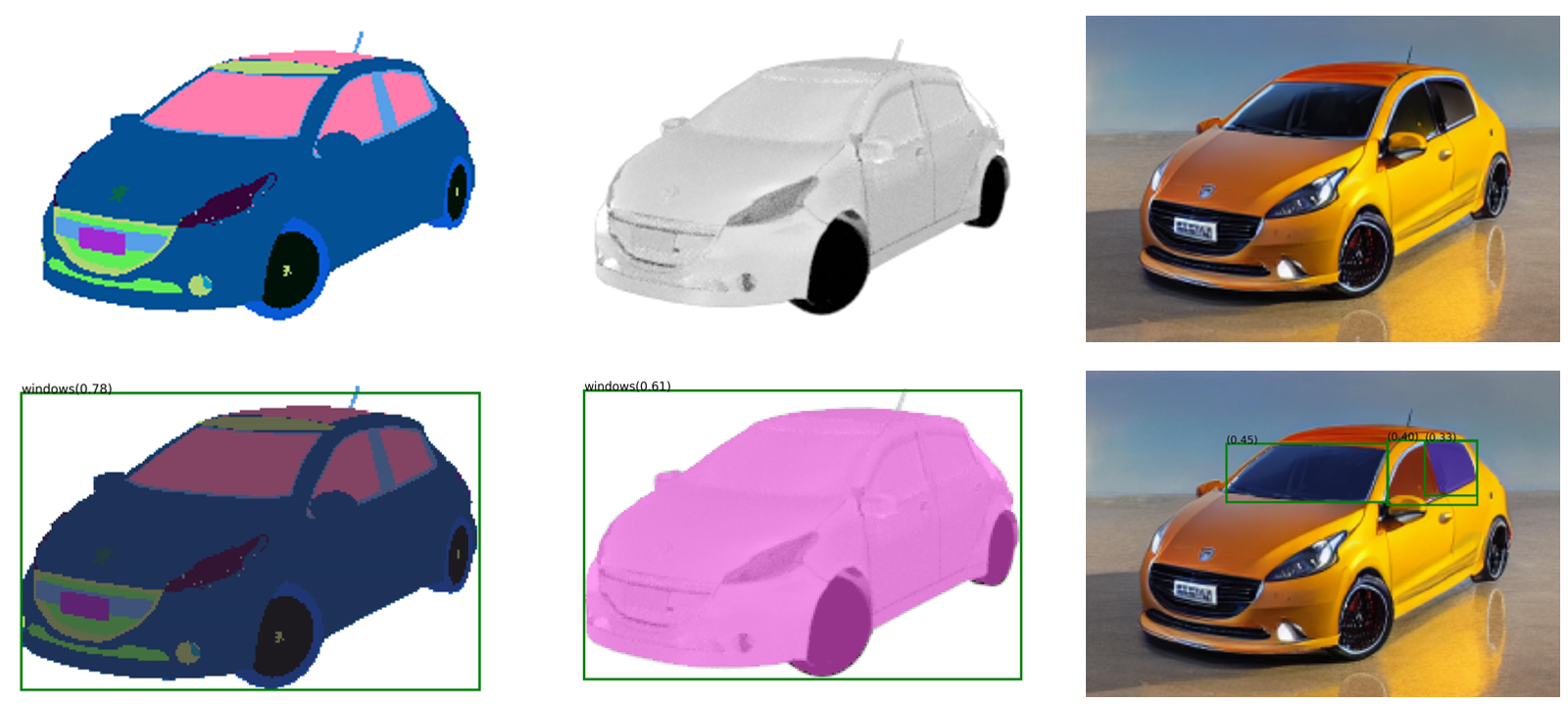}
    \vspace{-1em}
    \caption{\textbf{Window recognition results} on colored material design, retrieved CAD rendering, and augmented data by ControlNet~\cite{Zhang2023AddingCC}.}
    \vspace{-1em}
    \label{fig:recognition_illustration}
\end{figure}

\begin{figure}[htbp]
    \centering
    \includegraphics[width=0.9\linewidth]{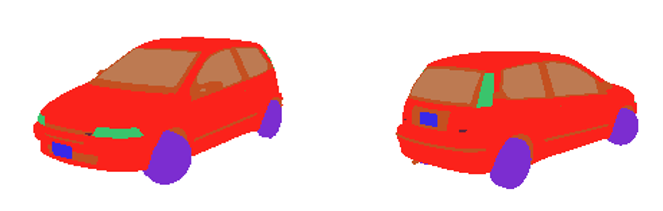}
    \vspace{-1em}
    \caption{\textbf{Symmetric material design}. Different colors represent different material indexes.}
    \vspace{-1em}
    \label{fig:mat_idx}
\end{figure}

\noindent \textbf{Semantic-based Part Material Prior Retrieval.} Considering that a single vehicle comprises various materials, we need to assign different parts of the CAD models with specific base procedural graphs. Importantly, we have obtained symmetric material design during our CAD retrieval, where disconnected components with the same semantic meaning (e.g., left and right windows) are assigned the same initial material index (refer to the supplementary \figref{fig:mat_idx}). The semantic meanings of these material indexes are unknown. Consequently, we only need to recognize the semantic meaning of the indexes in the material design for effective material retrieval. However, directly interpreting the semantic meanings from material designs or retrieved CAD model renderings can be inaccurate, as they often present unrealistic appearances (see \figref{fig:recognition_illustration}). To solve this, we use ControlNet~\cite{Zhang2023AddingCC} to produce photorealistic images based on the retrieved material design and use Grounded SAM ~\cite{Ren2024GroundedSA}, a foundational vision model that combines a large language model with image segmentation, to identify the part-level meanings of the retrieved material design (see the supplementary \figref{fig:part_ass} for illustration). To enhance the robustness further, we also implement multi-view recognition. 
For the car body, it is challenging to recognize it directly using text prompts. Therefore, after identifying the other components, we treat the largest area of the remaining part as the car body. We then retrieve base procedural graphs ($\b G_{init}$) based on the names of the recognized components. This method allows us to assign material priors to the entire vehicle robustly without the need for accurate part-segmentation.

\subsection{Material Optimization} \label{sec:optimization}

\noindent \textbf{Material Graph Differentiable Rendering.} 
We follow DiffMat v2~\cite{Li2023EndtoendPM} to convert the material node graph into texture elements like the albedo map $\bA_{uv}$, normal map $\b N_{uv}$, and roughness map $\bR_{uv}$. This produces a physically-based microfacet BRDF~\cite{karis2013real} model. To obtain the per-pixel material parameters $\bA$, $\b N$, and $\b R$, we use the UV sampling function $\b{Sample}$ to sample the material textures $\bA_{uv}$, $\b N_{uv}$, and $\bR_{uv}$ from the per-pixel texture (UV) coordinates $ \b{UV}$ of the UV map rendered in the matched pose (see supplementary \secref{sec:pose_matching}). Combined with the estimated incoming lighting $\b L$~\cite{Li2019InverseRF}, we perform differentiable rendering to achieve the estimated rendering results shown below:
\begin{align}
\b A_{uv}, \b N_{uv}, \b R_{uv} & = \b {DiffMat}(\b G) \\
\b A, \b N, \b R & = \b{Sample}(\b A_{uv}, \b N_{uv}, \b R_{uv}, \b{UV}) \\
\b I_{render} & = \b{Render}(\b A, \b N, \b R, \b L) 
\end{align}
where $\b{Render}$ is the differentiable renderer adopted from InvRenderNet~\cite{Li2019InverseRF}. 

\noindent \textbf{Part-Aware Material Optimization.} We also segment the components in the reference view using Grounded SAM and optimize the corresponding material of the CAD model to align with the reference view. Since there is no exact correspondence between rendered and reference pixels, we use a part-level loss $\b \ell_{stat}$ by minimizing the difference between the mean and variance of the corresponding parts following ~\cite{yeh2022photoscene}. To match the patterns of the reference view, we use a masked VGG loss $\ell_{vgg}$ using Gram matrices \cite{Gatys2016ImageST} to enhance visual similarity. To further align the color of the reference vehicles, we incorporate a masked RGB loss $\ell_{rgb}$ on the overlap region between components in the reference view and CAD model rendering. Please see the supplementary \secref{sec:mat_optimization_appendice} for details. The total loss optimizes the parameters of the material graphs with backpropagation. Note that optimizing complex material physical properties from single view images, such as opacity and roughness through differentiable rendering is not ideal. Instead, we only optimize the albedo of the retrieved metal and rubber materials to align with the input image and directly assign the retrieved glass material to the car window without further optimization, which can produce satisfactory results, as demonstrated in our experiments.

\section{Photorealistic Insertion in Urban Scenes} \label{sec:method2}

To construct photorealistic and controllable urban scenes with our optimized 3D CAD models, we need to seamlessly integrate them into the provided urban backgrounds.
Given multi-view background perspective and fisheye images, we begin by rendering the vehicles with lighting that matches the estimated environmental conditions (\secref{sec:env_map}). Next, we compose these rendered vehicles with the scene's background created through reconstruction methods (\secref{sec:blending}).

\subsection{Environment Lighting Estimation} \label{sec:env_map}

To render our vehicle models realistically, accurately estimating the scene's environment lighting map is essential. While per-pixel incoming lighting estimation (as discussed in \secref{sec:optimization}) is one option, it lacks global consistency and can cause artifacts when the vehicle is moved. To address this, we propose using multi-view fisheye images mounted on both sides of a car~\cite{liao2022kitti} to estimate the lighting environment. A pair of fisheye images provides a 360$^\circ$ field of view, enabling the construction of a globally consistent environment map for each pair. We first convert each fisheye pair into a panorama image. Given that the images captured by the fisheye cameras are in low dynamic range (LDR) format, we transform the upper half of the LDR panorama image into high dynamic range (HDR) to accurately represent the lighting of the skydome using a pre-trained network~\cite{Wei2024EditableSS}. Next, to incorporate other objects in the scene, we segment the non-sky region of the LDR panorama using FastSAM~\cite{Zhao2023FastSA} and then compose it with the HDR skydome after aligning the value ranges. This approach allows us to consider the surrounding lighting and shadow caused by foreground objects when rendering. To further achieve the spatially varying effect, we select the environment map closest to the insertion position based on the distances between the insertion position and the fisheye camera locations. Although this is a rough approximation compared to time-consuming ray-tracing techniques~\cite{Pun2023LightSimNL}, it results in a reasonable and robust performance.

\begin{figure*}[htbp]   
    \centering
    \def\mywidth{0.092\textwidth}
    \setlength{\tabcolsep}{0.8pt} %
    \begin{tabular}{P{0.3cm} >{\centering\arraybackslash}m{\mywidth} >{\centering\arraybackslash}m{\mywidth} >{\centering\arraybackslash}m{\mywidth} >{\centering\arraybackslash}m{\mywidth} >{\centering\arraybackslash}m{\mywidth} >{\centering\arraybackslash}m{\mywidth} >{\centering\arraybackslash}m{\mywidth} >{\centering\arraybackslash}m{\mywidth} >{\centering\arraybackslash}m{\mywidth} >{\centering\arraybackslash}m{\mywidth}}
    \rotatebox[origin=c]{90}{\small{Ref.}}    & 
        \includegraphics[width=0.08\textwidth]{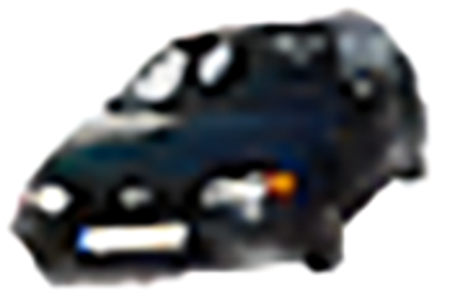} &  
        \includegraphics[width=0.08\textwidth]{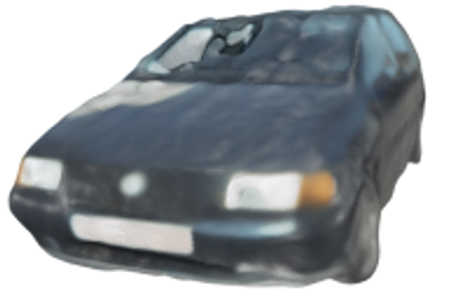} &  
        \includegraphics[width=0.08\textwidth]{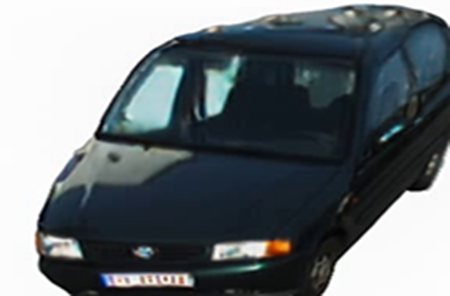} &  
        \includegraphics[width=0.08\textwidth]{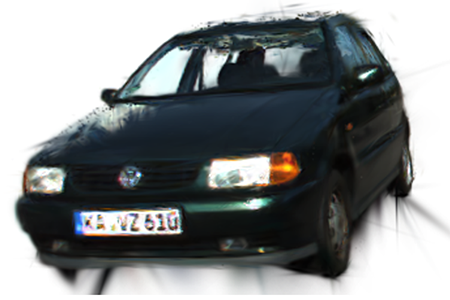} &  
        \includegraphics[width=0.08\textwidth]{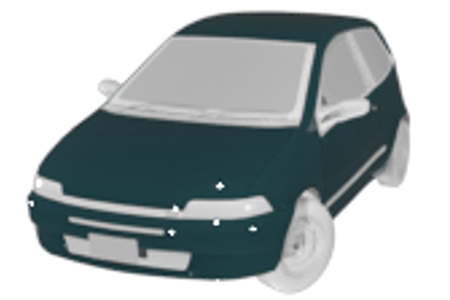} &  
        \includegraphics[width=0.08\textwidth]{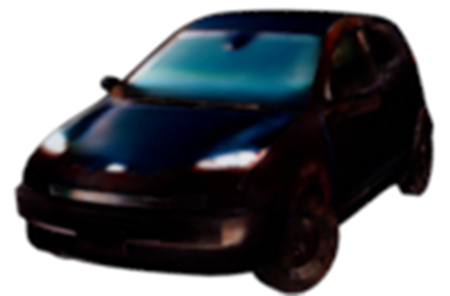} &  
        \includegraphics[width=0.08\textwidth]{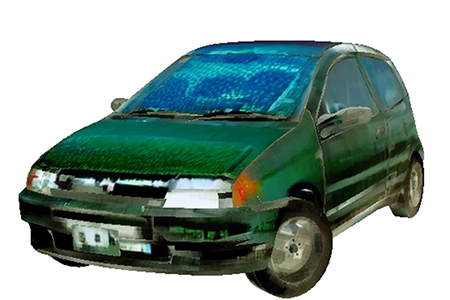} &  
        \includegraphics[width=0.08\textwidth]{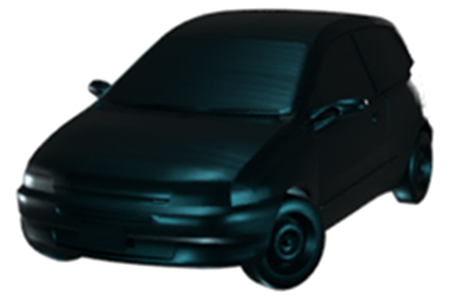} &
        \includegraphics[width=0.08\textwidth]{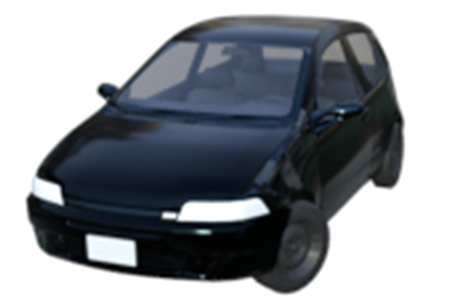} &
        \includegraphics[width=0.08\textwidth]{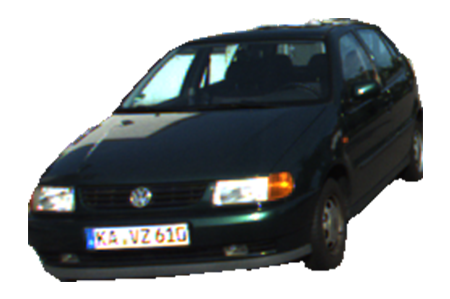}\\  
    \rotatebox[origin=c]{90}{\small{Rot.}}    &
        \includegraphics[width=0.08\textwidth]{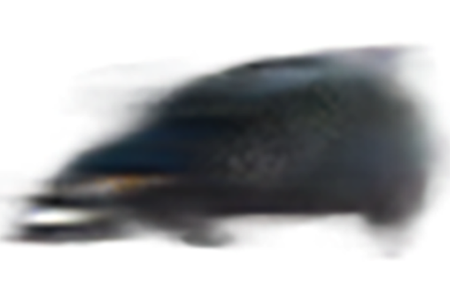} &  
        \includegraphics[width=0.08\textwidth]{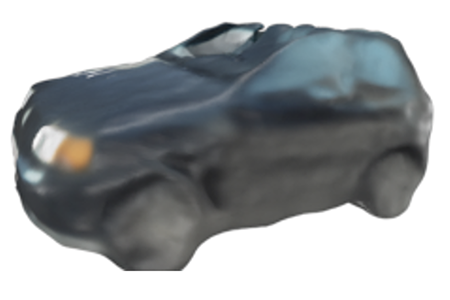} &  
        \includegraphics[width=0.08\textwidth]{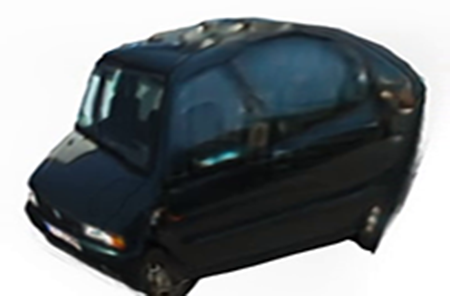} &  
        \includegraphics[width=0.08\textwidth]{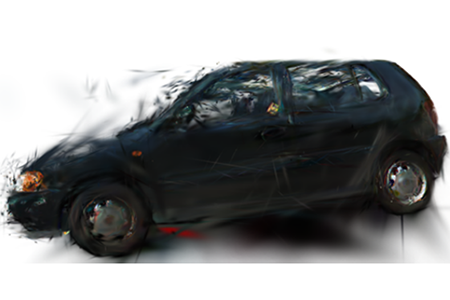} &  
        \includegraphics[width=0.08\textwidth]{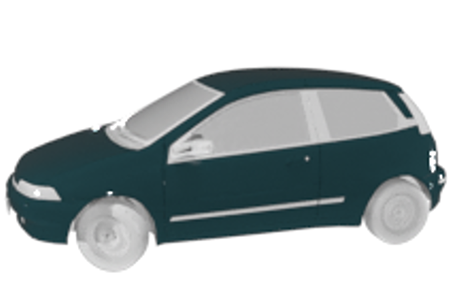} &  
        \includegraphics[width=0.08\textwidth]{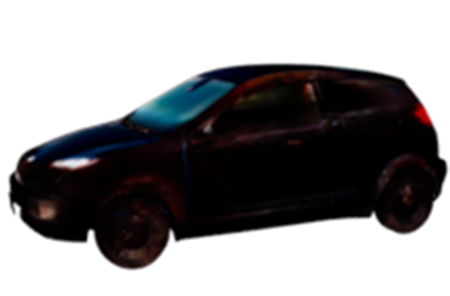} &  
        \includegraphics[width=0.08\textwidth]{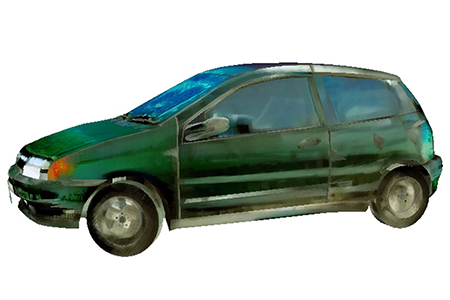} &
        \includegraphics[width=0.08\textwidth]{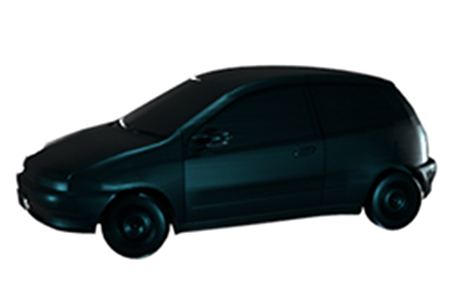} &
        \includegraphics[width=0.08\textwidth]{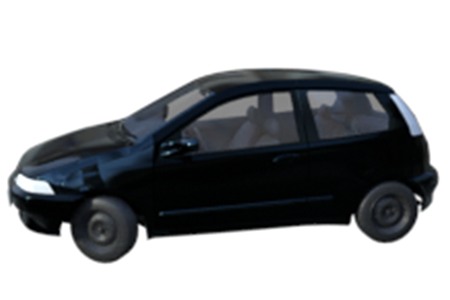} &
        \includegraphics[width=0.08\textwidth]{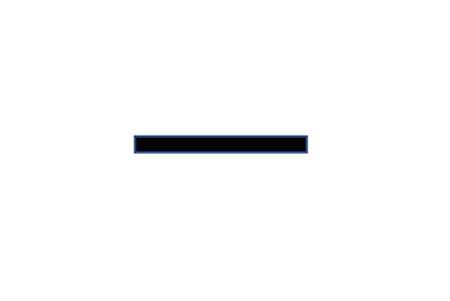}\\ 
    \rotatebox[origin=c]{90}{\small{Ref.}}    &
        \includegraphics[width=0.08\textwidth]{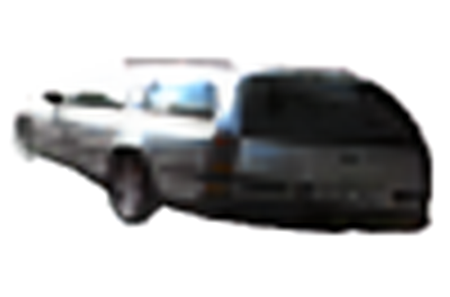} &  
        \includegraphics[width=0.08\textwidth]{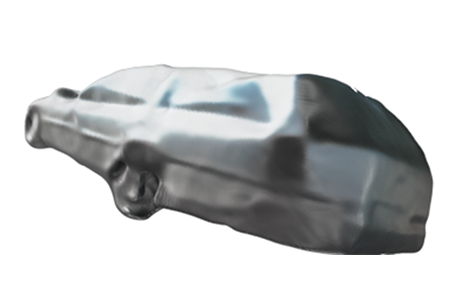} &  
        \includegraphics[width=0.08\textwidth]{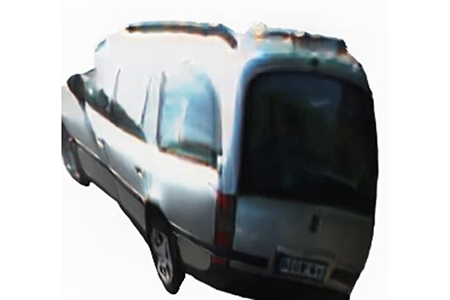} &  
        \includegraphics[width=0.08\textwidth]{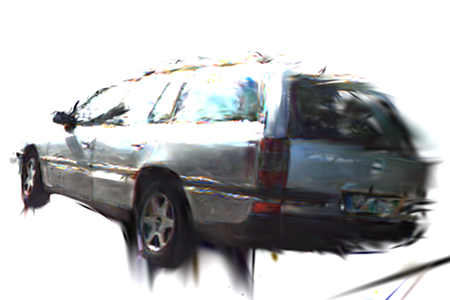} &  
        \includegraphics[width=0.08\textwidth]{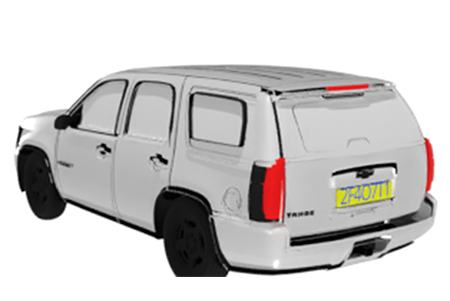} &  
        \includegraphics[width=0.08\textwidth]{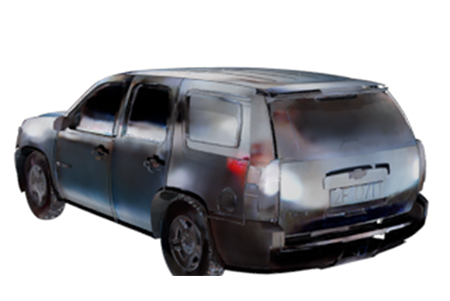} &  
        \includegraphics[width=0.08\textwidth]{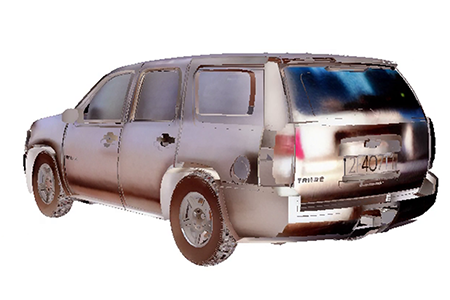} &
        \includegraphics[width=0.08\textwidth]{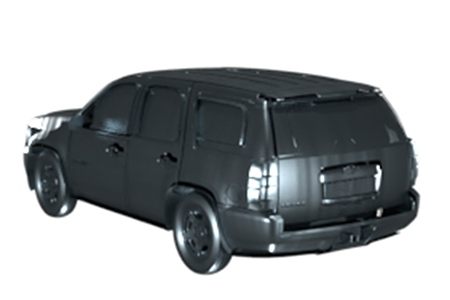} &
        \includegraphics[width=0.08\textwidth]{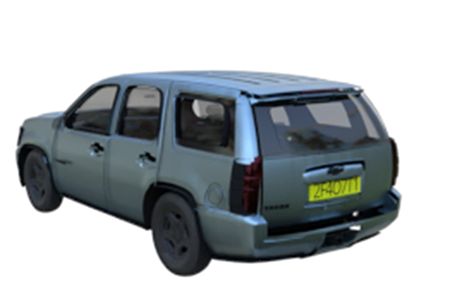} &
        \includegraphics[width=0.08\textwidth]{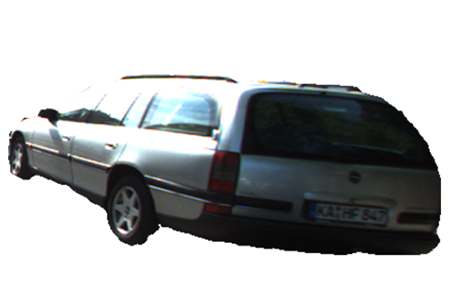}\\ 
    \rotatebox[origin=c]{90}{\small{Rot.}}    &
        \includegraphics[width=0.08\textwidth]{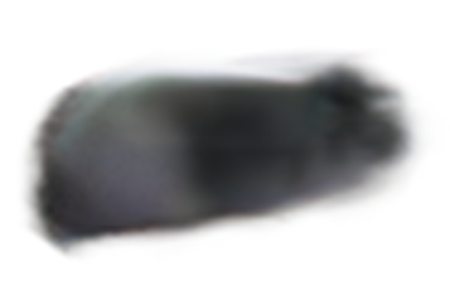} &  
        \includegraphics[width=0.08\textwidth]{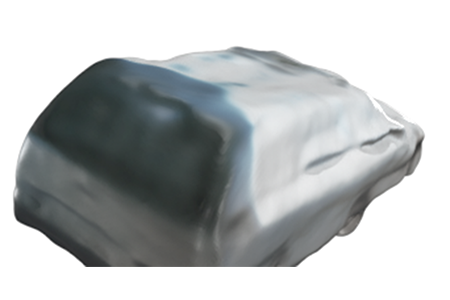} &  
        \includegraphics[width=0.08\textwidth]{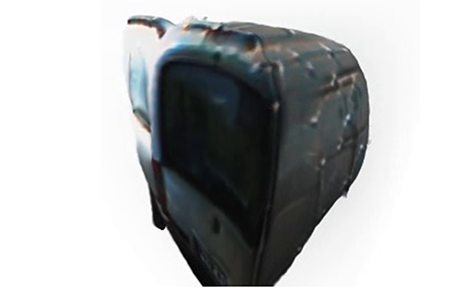} &  
        \includegraphics[width=0.08\textwidth]{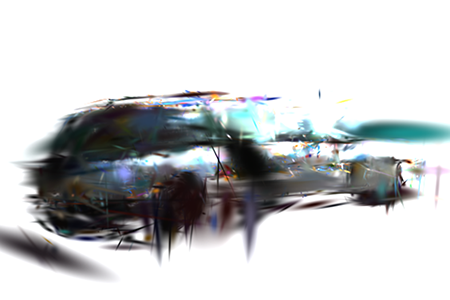} &  
        \includegraphics[width=0.08\textwidth]{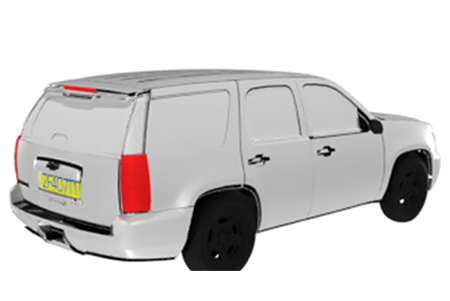} &  
        \includegraphics[width=0.08\textwidth]{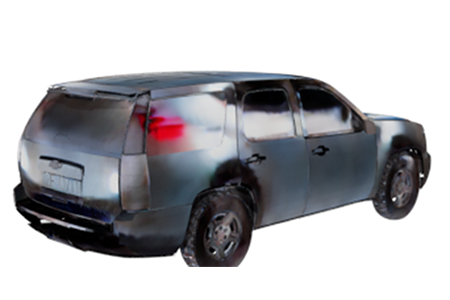} &  
        \includegraphics[width=0.08\textwidth]{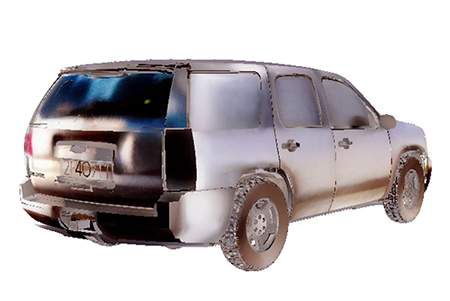} &
        \includegraphics[width=0.08\textwidth]{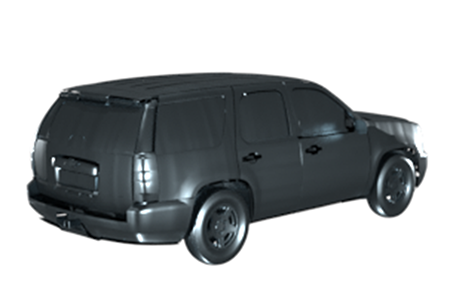} &
        \includegraphics[width=0.08\textwidth]{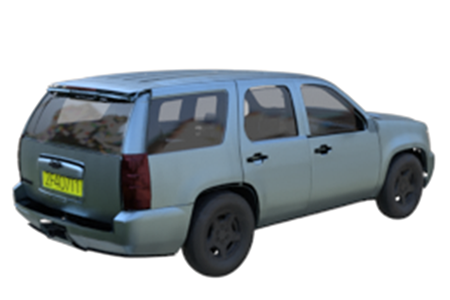} &
        \includegraphics[width=0.08\textwidth]{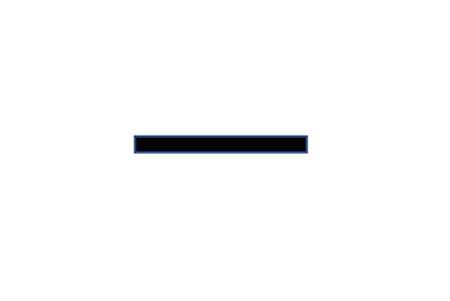}\\
        \vspace{-1ex} 
        & \scriptsize{PixNeRF~\cite{Yu2020pixelNeRFNR}} & \scriptsize{Wonder3D~\cite{Long2023Wonder3DSI}} &
        \scriptsize{LRM~\cite{hong2023lrm}} &
        \scriptsize{HUGS~\cite{zhou2024hugs}} & 
        \scriptsize{Ours (w/o opt.)} & \scriptsize{LatentPaint$^{\dag}$~\cite{Metzer2022LatentNeRFFS}} &
        \scriptsize{Paint3D$^{\dag}$~\cite{Zeng2023Paint3DPA}} &
        \scriptsize{PhotoScene$^{\dag}$\cite{yeh2022photoscene}} & 
        \scriptsize{Ours} & 
        \scriptsize{GT}\\
    \end{tabular}  
    \vspace{-0.5em}
    \caption{\textbf{Qualitative results} on KITTI-360 for novel view synthesis from reference (Ref.) and rotated (rot.) viewpoints. UrbanCAD produces more robust and realistic results at the novel viewpoint compared to the baselines.}  
    \vspace{-1em}
    \label{fig:app_comparison}  
\end{figure*}

\subsection{Background Reconstruction and Compostion} \label{sec:blending}

Since autonomous driving simulators require free navigation within the scene, we integrate our method with novel view synthesis (NVS) to enable this functionality. Given input background videos, we use the 3D Gaussian Splatting method ~\cite{zhou2024hugs} to reconstruct the environment and render background images from novel views. Subsequently, we render the foreground vehicle using Blender~\cite{blender} and blend it with the background images via alpha composition. Specifically, we position the vehicle models generated in \secref{sec:method1} in Blender according to a target 3D position obtained from annotations or trajectory generation methods~\cite{Zhang2023CATCA}. A virtual plane is utilized to account for shadow effects based on the estimated ground plane. Using the previously estimated HDR environment map, we render the vehicles and composite the rendered vehicles with the background images. Please refer to the supplementary \secref{sec:background_rec} for details.

\section{Experiment} \label{sec:experiment}

\subsection{Experiment Setup}

\noindent \textbf{Datasets.} We evaluate our method on various urban datasets. 
We mainly conduct experiments on the KITTI-360 dataset \cite{liao2022kitti} which contains high-quality fisheye images. However, we also present CAD model optimization results on the Multi-View Marketplace Cars (MVMC)~\cite{Zhang2021NeRSNR} dataset since our CAD retrieval and optimization module can function without fisheye images.
For the CAD models, we utilize the Objaverse library \cite{deitke2023objaverse}, a free 3D asset repository containing 26k+ car and vehicle models. For the base procedural material graphs, we collect them from the Adobe 3D Asset Library~\cite{Adobe} containing 13k+ materials.

\noindent \textbf{Baselines.} We compare our approach with various types of methods: (1) Single-view reconstruction method using the conditional implicit function: PixelNeRF \cite{Yu2020pixelNeRFNR}. (2) Single-view generation method using diffusion prior: Wonder3D \cite{Long2023Wonder3DSI}. (3) Single-view reconstruction method using large reconstruction model: LRM~\cite{hong2023lrm}. (4) Multi-view reconstruction method using 3DGS: HUGS \cite{zhou2024hugs}. 
(5) Mesh texturing methods using the generative model: LatentPaint$^{\dag}$~\cite{Metzer2022LatentNeRFFS} and Paint3D$^{\dag}$~\cite{Zeng2023Paint3DPA}. (6) Mesh texturing method using optimizable procedural graph: PhotoScene$^{\dag}$~\cite{yeh2022photoscene}. Note that we use our CAD retrieval module (\secref{sec:retrieval}) to retrieve CAD models before mesh texturing with LatentPaint~\cite{Metzer2022LatentNeRFFS}, Paint3D~\cite{Zeng2023Paint3DPA} and PhotoScene~\cite{yeh2022photoscene} (marked as ``LatentPaint$^{\dag}$", ``Paint3D$^{\dag}$" and ``PhotoScene$^{\dag}$"). We also investigate the performance when directly using our CAD retrieval module without material retrieval and optimization (marked as UrbanCAD (w/o opt.) ).

\noindent \textbf{Metrics.} Given our focus on the controllable aspects of digital twins, we manipulate the vehicles to be rendered from different viewpoints and evaluate the Fréchet Inception Distance (FID) \cite{heusel2017gans} and Kernel Inception Distance (KID) \cite{binkowski2018demystifying} between these renderings and real-world car datasets~\cite{Yang2015ALC}. This assesses the photorealism in terms of free viewpoint controllability. To evaluate the reconstruction quality, we also calculate the Learned Perceptual Image Patch Similarity (LPIPS)~\cite{Zhang2018TheUE} between the input reference vehicle images and the renderings under matched poses (detailed in the supplementary). Additionally, we evaluate the performance of self-driving perception methods on our generated synthetic data using Intersection over Union (IOU) and Panoptic Quality (PQ) \cite{Kirillov2018PanopticS} metrics for all vehicles in the scenes. We also assess corner cases using both category-level IOU for all vehicles and instance-level IOU for a specific vehicle.

\subsection{Photorealism Quality}
\begin{figure*}[htbp]   
    \centering  
    \def\mywidth{0.1\textwidth}
    \setlength{\tabcolsep}{1pt} %
    \begin{tabular}{cccccccc}
        \includegraphics[width=0.12\textwidth]{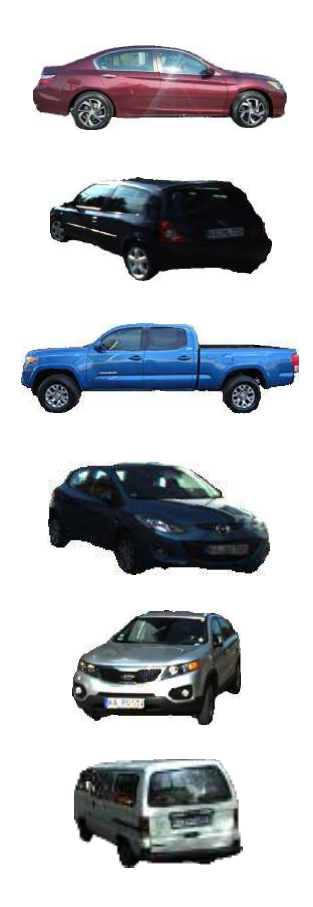} &  
        \includegraphics[width=0.12\textwidth]{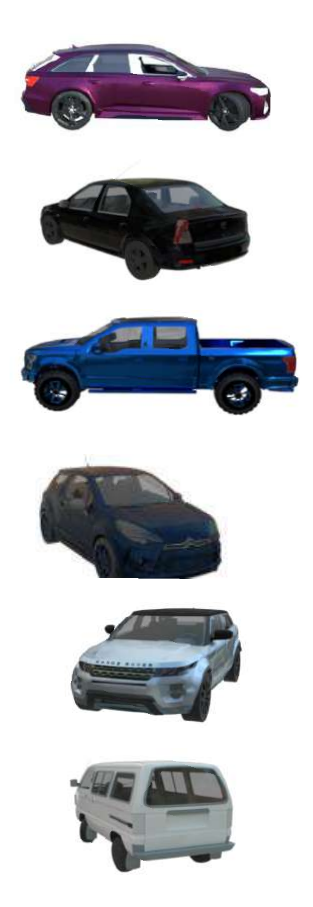} &  
        \includegraphics[width=0.12\textwidth]{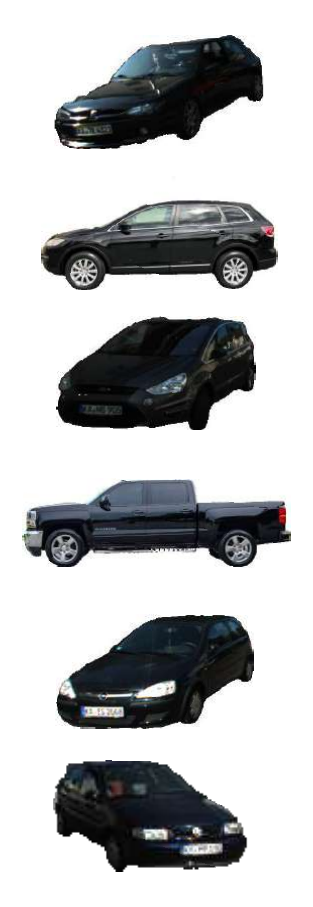} &  
        \includegraphics[width=0.12\textwidth]{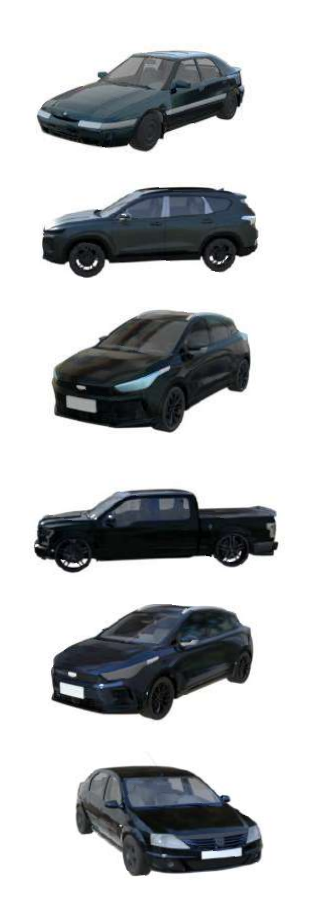} &  
        \includegraphics[width=0.12\textwidth]{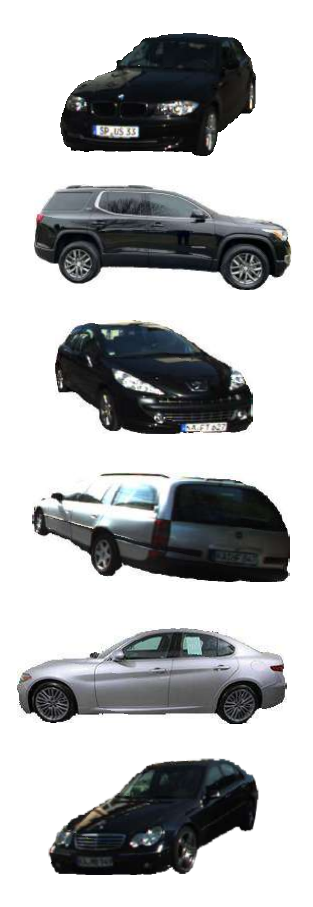} &  
        \includegraphics[width=0.12\textwidth]{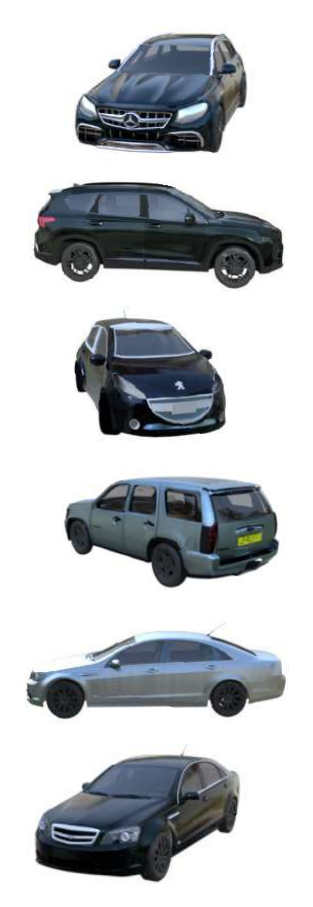} &  
        \includegraphics[width=0.12\textwidth]{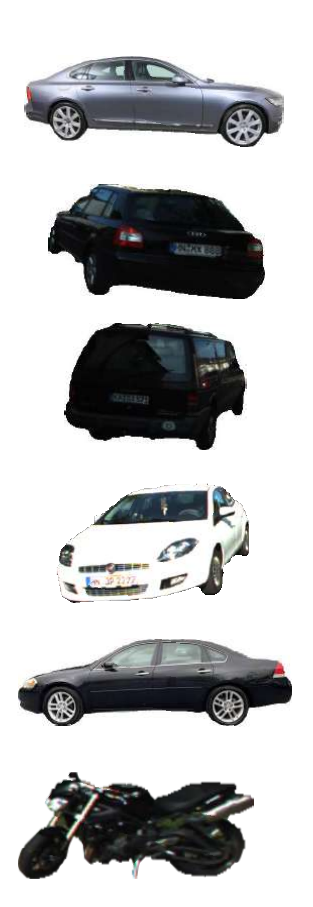} &  
        \includegraphics[width=0.12\textwidth]{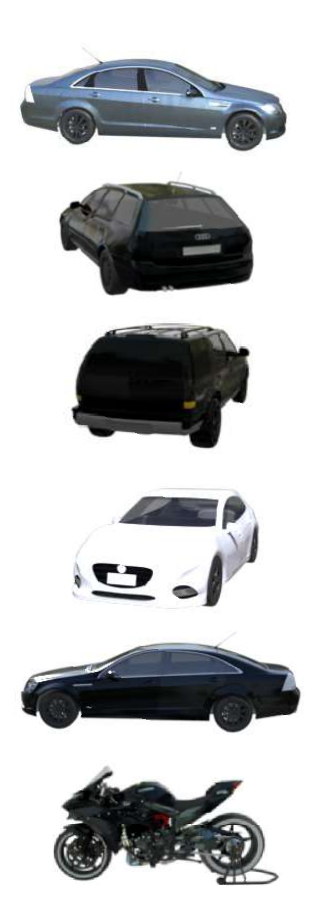} \\
        \vspace{-1em}
        \footnotesize{Ref.} & \footnotesize{Ours} &
        \footnotesize{Ref.} & \footnotesize{Ours} & 
        \footnotesize{Ref.} & \footnotesize{Ours} &
        \footnotesize{Ref.} & \footnotesize{Ours}\\
    \end{tabular}  
    \caption{\textbf{More pairs} of 3D vehicles after CAD model retrieval and material optimization (right) alongside the input single-view segmented vehicles (left). UrbanCAD produces photorealistic 3D vehicles with different categories given single-view inputs.}  
    \vspace{-1em}
    \label{fig:object_pairs}  
\end{figure*}

\begin{table}[t]
\centering
\resizebox{.49\textwidth}{!} 
{
\begin{tabular}{cc|c|ccc}
\toprule
 Reconstruction-based & Retrieval-based & Method & FID$\downarrow$ & KID$\downarrow$ & LPIPS$\downarrow$\\
\midrule
 \checkmark & & PixelNeRF~\cite{Yu2020pixelNeRFNR} & 264.61 & 0.2415 & \--{}\\
 \checkmark & & Wonder3D~\cite{Long2023Wonder3DSI} & 246.43 & 0.2292 & \--{}\\
 \checkmark & & LRM~\cite{hong2023lrm} & 220.77 & 0.2050 & \--{}\\
 \checkmark & & HUGS~\cite{zhou2024hugs} & 240.92 & 0.2417 & \--{}\\
  & \checkmark & UrbanCAD (w/o opt.) & 81.05 & 0.0567 & 0.6174\\
  & \checkmark & LatentPaint$^{\dag}$~\cite{Metzer2022LatentNeRFFS} & 85.62 & 0.0604 & 0.5525\\
  & \checkmark & Paint3D$^{\dag}$~\cite{Zeng2023Paint3DPA} & 67.52 & \textbf{0.0417} & 0.5652\\
  & \checkmark & PhotoScene$^{\dag}$~\cite{yeh2022photoscene} & 170.21 & 0.1561 & 0.5422\\
  & \checkmark & UrbanCAD (Ours) & \textbf{62.80} & 0.0479 & \textbf{0.5242}\\
\bottomrule
\end{tabular}
}
\vspace{-0.5em}
\caption{\textbf{Quantitative Comparison} on the photorealism.}
\vspace{-1.5em}
\label{tab:app_comparison}

\end{table}

\noindent \textbf{Comparison to baselines.} As shown in \figref{fig:teaser}, UrbanCAD successfully reconstructs photorealistic vehicles within the provided urban images. We compare our method with baseline approaches both qualitatively (\figref{fig:app_comparison}) and quantitatively (\tabref{tab:app_comparison}). Given that self-driving simulation systems require vehicles to move freely within the scene, we focus on novel view rendering results across 360$^\circ$. We report the FID and KID metrics for the 360$^\circ$ renderings of the vehicles in \tabref{tab:app_comparison}. UrbanCAD demonstrates superior performance on FID and KID compared to most baselines, indicating that our vehicle models are more realistic. Interestingly, even only using our CAD retrieval module (UrbanCAD (w/o opt.)) outperforms the other reconstruction-based baselines in terms of FID and KID. This can be attributed to the fact that the reconstruction-based baselines may only provide reasonable reconstruction near the reference viewpoint (PixelNeRF, HUGS, Wonder3D, LRM). We also report the LPIPS metrics between reference images and CAD model renderings under matched poses for retrieval-based methods in \tabref{tab:app_comparison}. Compared to the baselines, UrbanCAD has better performance on LPIPS, suggesting that our vehicle models are more similar to the vehicles in the reference image. This is probably because the baselines lack high-frequency details and accurate material estimation (LatentPaint$^{\dag}$, PhotoScene$^{\dag}$, Paint3D$^{\dag}$). Our qualitative results in \figref{fig:app_comparison} and \figref{fig:more_app_comparison} further demonstrate that UrbanCAD produces superior outcomes, especially at large rotation angles. Besides, we provide a quantitative comparison with a 3D reconstruction method ~\cite{Zhang2021NeRSNR} based on a surface implicit model (see the supplementary \tabref{tab:ners}). We also notice that Paint3D$^{\dag}$ performs better on KID compared to our method. This is probably because Paint3D tends to generate unrealistic yet rich textures as shown in \figref{fig:app_comparison} and \figref{fig:more_app_comparison}.

\noindent \textbf{Generalization ability.} Thanks to the large scale of the CAD model library and our material optimization module, our method exhibits strong generalization ability on various kinds of vehicles including cars, trucks, vans, and motorcycles as shown in \figref{fig:object_pairs}.

\noindent \textbf{Lighting estimation results.} We compare our fisheye-based lighting estimation module with baselines qualitatively (see the supplementary \secref{sec:lighting_comp}). The results demonstrate that our method can produce more accurate lighting for photorealistic vehicle insertion in urban scenes.

\noindent \textbf{Ablation study.} The comparison against UrbanCAD (w/o opt.) and PhotoScene$^{\dag}$ in \figref{fig:app_comparison} and \tabref{tab:app_comparison} highlights the significance of our retrieval-based material optimization and part-aware material prior retrieval modules. Results in \tabref{tab:self-driving} and supplementary \secref{sec:lighting_comp} demonstrate our lighting estimation module is crucial for constructing photorealistic scenarios.

\begin{figure}[h]
    \centering
    \small
    \setlength{\tabcolsep}{0pt}
    \includegraphics[width=\linewidth]{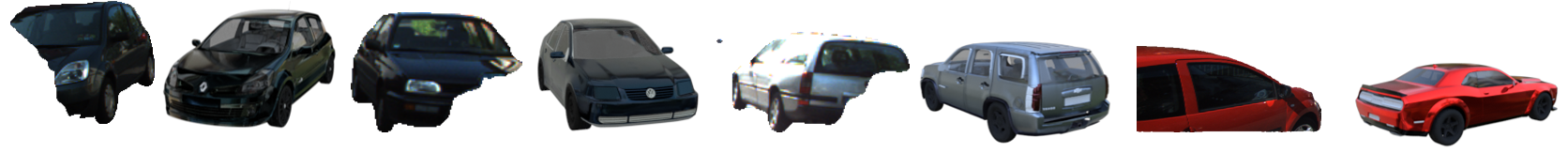}
    \vspace{-0.5em}
        \begin{tabularx}{\linewidth}{XXXXXXXX}
        \centering
          \footnotesize{Ref.} &  \footnotesize{Ours.} & \footnotesize{Ref.} &  \footnotesize{Ours.} & \footnotesize{Ref.} &  \footnotesize{Ours.} & \footnotesize{Ref.} &  \footnotesize{Ours.} \\
        \end{tabularx}
    \caption{Recovery from partial observation.}
    \vspace{-1.5em}
    \label{fig:occlusion}
\end{figure}

\noindent \textbf{Robustness to occlusion.} We present more results based on partial observations in \figref{fig:occlusion}. We find SAM is robust to occlusion, likely due to its training on occluded images. Our CAD retrieval module demonstrates resilience to occlusion as well, as it utilizes a 3D encoder aligned with CLIP's latent space to encode CAD models and retrieve them based on semantic instead of visual similarity. In addition, our material optimization module produces reliable results by leveraging part-aware regularization between corresponding semantic regions, eliminating the need for precise pixel-level alignment.

\subsection{Downstream Application.} \label{sec:downstream}

\begin{table*}[t]
\centering
\resizebox{.95\textwidth}{!} 
{
\begin{tabular}{cc|cc|cc|cc|cc}
\toprule
& & \multicolumn{2}{c|}{YOLO-Seg} & \multicolumn{2}{c|}{Mask2Former (R50)}& \multicolumn{2}{c|}{Mask2Former (R101)}  & \multicolumn{2}{c}{Mask2Former (SwinL)}\\
\multicolumn{2}{c|}{Data} & IOU$\uparrow$ & PQ$\uparrow$ & IOU$\uparrow$ & PQ$\uparrow$& IOU$\uparrow$ & PQ$\uparrow$ & IOU$\uparrow$& PQ $\uparrow$\\
\midrule
\multicolumn{2}{c|}{Real-world data} & 68.27 & 62.91 & 66.15 & 38.20 & 73.04 & 32.05 & 86.89 & 48.99\\
\midrule
\midrule
& Trajectory 1 & 70.66 & \textbf{62.60} & 61.85 & 28.40 & 71.32 & \underline{41.50} & 73.15 & 47.11\\
Wonder3D & Trajectory 2 & 71.06 & \textbf{63.12} & 62.58 & 28.48 & 71.66 & 40.41 & 73.14 & 46.87\\
& Trajectory 3 & 71.08 & \textbf{62.78} & 63.60 & 29.34 & 72.04 & 40.62 & 73.08 & 47.74\\
\midrule
& Trajectory 1 & 69.21 & 60.06 & \underline{67.28} & 32.61 & 72.89 & 40.13 &  \textbf{84.33} & \textbf{52.66}\\
UrbanCAD (w/o material optimization) & Trajectory 2 & 68.69 & 59.88 & \underline{67.10} & 32.01 & 73.14 & 39.95 & \textbf{84.89} & \textbf{52.54}\\
& Trajectory 3 & 67.06  & 59.07 & \underline{67.53} & 32.10 & 73.42 & 40.62 & \textbf{84.63} & \textbf{52.67}\\
\midrule
& Trajectory 1 & \underline{73.11} & 61.19 & 65.45 & \underline{33.17} & \underline{74.09} & 41.10 & 82.57 & 51.34\\
UrbanCAD (w/o lighting estimation)& Trajectory 2 & \underline{73.33} & 61.30 & 65.46 & \underline{32.99} & \underline{74.65} & \underline{41.13} & 83.34 & 51.27\\
& Trajectory 3 & \underline{73.22} & 60.67 & 65.57 & \underline{32.89} & \underline{73.84} & \underline{41.78} & 83.28 & 51.33\\
\midrule
& Trajectory 1 & \textbf{74.08} & \underline{61.60} & \textbf{70.28} & \textbf{34.71} & \textbf{76.40} & \textbf{43.47} & \underline{83.83} & \underline{52.27}\\
UrbanCAD (Ours) & Trajectory 2 & \textbf{74.15} & \underline{62.16} & \textbf{70.07} & \textbf{34.10} & \textbf{76.91} & \textbf{43.56} & \underline{84.45} & \underline{52.12}\\
& Trajectory 3 & \textbf{74.02} & \underline{61.64} & \textbf{70.42} & \textbf{34.45} & \textbf{76.56} & \textbf{43.55} & \underline{84.23} & \underline{52.39} \\
\bottomrule
\end{tabular}
}
\captionof{table}{\textbf{Quantitative Comparison} of perception methods on different data. \underline{Underline} denotes second best.}
\vspace{-1em}
\label{tab:self-driving}

\end{table*}

To further explore whether our optimized CAD models facilitate the rendering of photorealistic images that can enhance downstream applications, we evaluate several pre-trained segmentation models on our augmented data, which consists of optimized CAD models blended with ground truth (GT) images. Specifically, we test the pre-trained YOLO V8 instance segmentation model, a widely used real-time instance segmentation method that combines the YOLO V8 detection model \cite{Reis2023RealTimeFO} with YOLACT \cite{Bolya2019YOLACTRI}. Additionally, we assess the performance of the instance segmentation model Mask2Former~\cite{Cheng2021MaskedattentionMT}, utilizing different backbones, on both our synthetic data and real-world reference data.

\noindent \textbf{In-Distribution Driving Scenarios.} Firstly, we construct normal in-distribution driving scenarios using the vehicles generated by Wonder3D, UrbanCAD without retrieval-based material optimization, UrbanCAD without lighting estimation, and UrbanCAD in full setting, to evaluate the perception model's performance. Please see the supplementary \secref{sec:syn_data} for details.
We also select 100 frames of real-world images with similar vehicle distribution and positions compared to our synthetic scenes. 
\tabref{tab:self-driving} shows that the YOLO-seg model and Mask2Former with ResNet~\cite{He2015DeepRL} backbones achieve better IOU results on synthetic scenarios constructed with the UrbanCAD (ours) vehicle models. Additionally, Mask2Former with ResNet backbones reports a higher PQ value on scenarios constructed by UrbanCAD (ours). These results demonstrate that UrbanCAD (ours) can produce high-quality synthetic data for perception tasks. Interestingly, we find that both material optimization and lighting estimation are essential for constructing synthetic scenarios with a small domain gap. The performance of perception models degrades when these modules are removed. 
The small drop in PQ value compared to Wonder3D is due to the different geometries between Wonder3D vehicles and UrbanCAD (ours) vehicles, which leads to different GT values. We observe that the Mask2Former model using the large Swin Transformer backbone~\cite{Liu2021SwinTH} exhibits strong generalization ability. However, inference speed is crucial for self-driving applications, and the Mask2Former model with the SwinL backbone is limited by low inference speed. Furthermore, performance on real-world data is notably lower because exact real-world data corresponding to our synthetic scenarios are not obtainable, as we modify the ground truth of the background images when inserting the vehicles.

\noindent \textbf{Out-of-Distribution Driving Scenarios.} Thanks to the high controllability and photorealism of our generated vehicle models, we demonstrate that our method can generate photorealistic corner cases, as shown in \figref{fig:teaser} and supplementary \figref{fig:components editing}, to challenge existing perception models. Since measuring perception results based on windows and tires in safety-critical scenarios is difficult, we focus on the performance of the perception system in door-opening settings. Specifically, we construct five door opening and closing scenarios with a total of 150 frames and test the perception system on these corner-case scenarios and reference images where vehicle doors remain closed. Given that opening and closing vehicle doors lead to small changes in the overall scene, we also report the instance-level IOU. As shown in \tabref{tab:corner-case}, the performance of the self-driving perception system declines rapidly in such out-of-distribution scenarios, despite the same perception model performing well in the in-distribution setting as shown in \tabref{tab:self-driving}. 
This underscores the importance of constructing urban scenarios with our highly controllable vehicles to test self-driving perception systems.

\begin{table}[t]
\centering
\tiny
\resizebox{.49\textwidth}{!} 
{
\setlength{\tabcolsep}{2pt} 
\begin{tabular}{c|cc|cc|cc}
\toprule
& \multicolumn{2}{c|}{Yolo-Seg} & \multicolumn{2}{c|}{Mask2Former (R101)}  & \multicolumn{2}{c}{Mask2Former (SwinL)}\\
Data & IOU$\uparrow$ & iIOU$\uparrow$ & IOU$\uparrow$ & iIOU$\uparrow$ & IOU$\uparrow$& iIOU $\uparrow$\\
\midrule
Reference data & 75.76 & \textbf{81.06} & 88.89 & \textbf{96.47} & 91.36 & \textbf{95.67}\\
OOD data (Ours) & 75.08 & 72.20 & 87.15 & 83.46 & 90.93 & 85.44\\
\bottomrule
\end{tabular}
}
\vspace{-1em}
\captionof{table}{\textbf{Quantitative Comparison} on reference data and out-of-distribution data generated by UrbanCAD.}
\vspace{-3em}
\label{tab:corner-case}

\end{table}

\vspace{-1em}

\section{Conclusion and Limitations} \label{sec:con}

In this paper, we aim to create photorealistic and highly controllable 3D vehicle digital twins for constructing challenging and realistic scenarios. Towards this goal, we introduce UrbanCAD, a framework that generates 3D vehicle digital twins with photorealistic appearances and high controllability through CAD model retrieval and optimization. Additionally, by reconstructing the background and environmental lighting, UrbanCAD facilitates the realistic insertion of our generated vehicle models into urban scenes. We demonstrate UrbanCAD's capabilities in producing photorealistic and highly controllable 3D vehicle digital twins, as well as in creating realistic, challenging, and safety-critical scenarios to test the robustness of self-driving perception systems. However, due to the semantically aligned CAD retrieval, the geometries of our created CAD models are not the same as the vehicles in the input image. Besides, our estimated spatially varying environment lighting may be not accurate if the insertion position is far from our fisheye cameras.

{
    \small
    \bibliographystyle{ieeenat_fullname}
    \bibliography{bibliography, bibliography_long, bibliography_custom}
}

\clearpage
\setcounter{page}{1}
\maketitlesupplementary

This appendix details our method, implementation, experimental designs, additional experiment results, utilized resources, and broader implications. We first detail how to retrieve and optimize the CAD models in \secref{sec:cad_filtering}, \secref{sec:pose_matching}, \secref{sec:mat_prior_retrieval}, \secref{sec:car_body_merging}, and \secref{sec:car_body_merging}, and then we show the process of urban lighting estimation in \secref{sec:lighting} and background reconstruction in \secref{sec:background_rec}. In \secref{sec:exp_detail}, we provide details on experiment designs including baselines implementation (\secref{sec:base_imp}), synthetic data generation (\secref{sec:syn_data}), and perception systems implementation (\secref{sec:per_imp}). We also show more results and implementation details of our functionality in (\secref{sec:func}). Finally, we report additional experiments and analysis in (\secref{sec:more}).

\section{ UrbanCAD Implementation Details} \label{sec:sup_imp}

\subsection{ CAD Model Filtering } \label{sec:cad_filtering}

Our method requires the CAD models to have correct material index assignment to support automatic coloring. However, we observe that in free CAD model libraries, there are small parts of handcrafted CAD models without proper material index designs. To this end, we design a script to filter the unqualified CAD models automatically or with a small amount of user interface based on the material design.

\begin{figure*}[htbp]   
    \centering  
    \def\mywidth{0.1\textwidth}
    \setlength{\tabcolsep}{1pt} %
    \begin{tabular}{P{0.3cm} >{\centering\arraybackslash}m{\mywidth} >{\centering\arraybackslash}m{\mywidth} >{\centering\arraybackslash}m{\mywidth} >{\centering\arraybackslash}m{\mywidth} >{\centering\arraybackslash}m{\mywidth} >{\centering\arraybackslash}m{\mywidth} >{\centering\arraybackslash}m{\mywidth} >{\centering\arraybackslash}m{\mywidth} >{\centering\arraybackslash}m{\mywidth}}
    \rotatebox[origin=c]{90}{\small{Ref.}}    & 
        \includegraphics[width=0.1\textwidth]{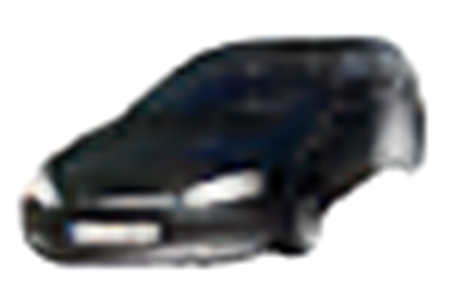} &  
        \includegraphics[width=0.1\textwidth]{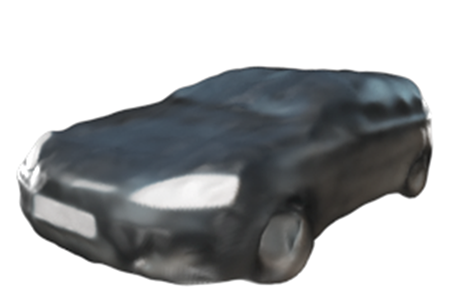} &  
        \includegraphics[width=0.1\textwidth]{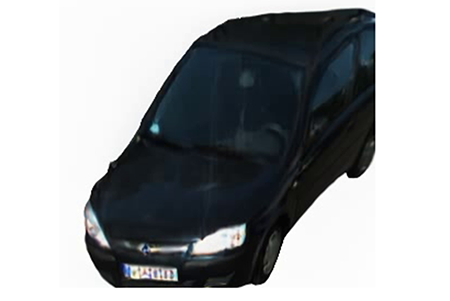} &  
        \includegraphics[width=0.1\textwidth]{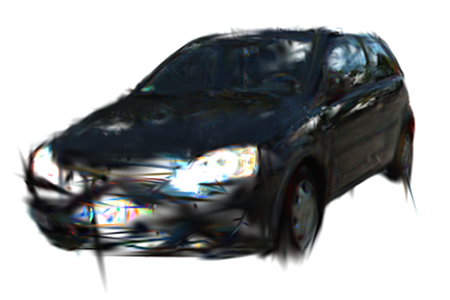} &  
        \includegraphics[width=0.1\textwidth]{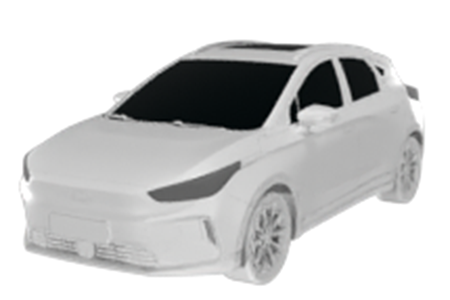} &  
        \includegraphics[width=0.1\textwidth]{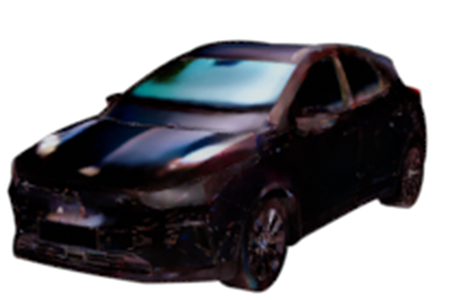} &  
        \includegraphics[width=0.1\textwidth]{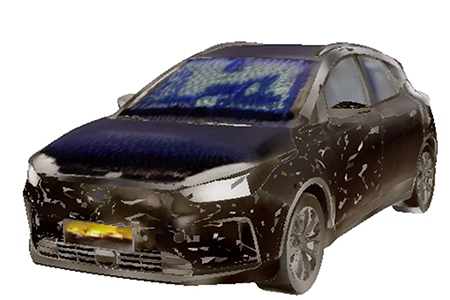} &  
        \includegraphics[width=0.1\textwidth]{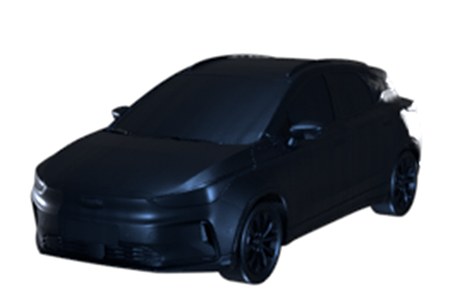} &  
        \includegraphics[width=0.1\textwidth]{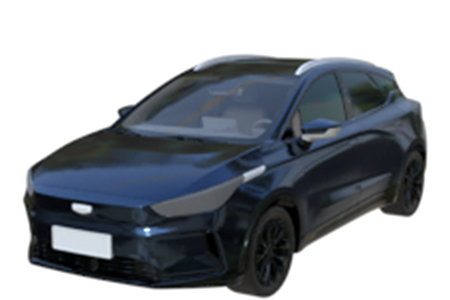}\\  
    \rotatebox[origin=c]{90}{\small{Rot.}}    &
        \includegraphics[width=0.1\textwidth]{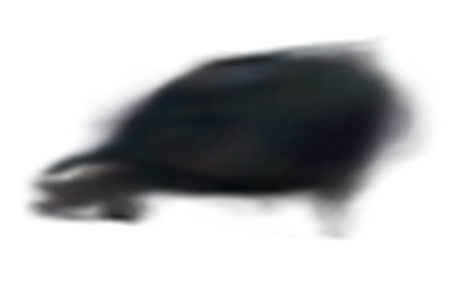} &  
        \includegraphics[width=0.1\textwidth]{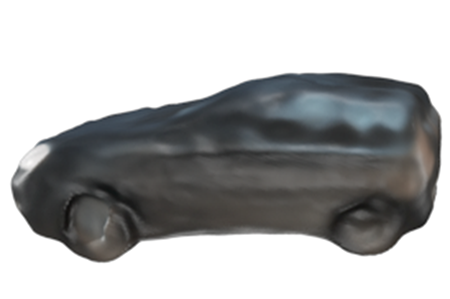} &  
        \includegraphics[width=0.1\textwidth]{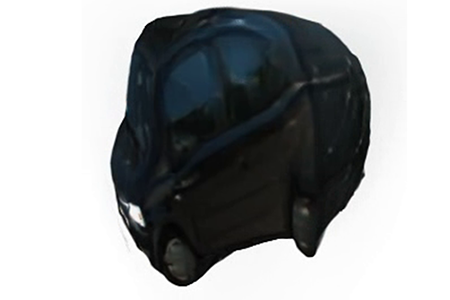} &  
        \includegraphics[width=0.1\textwidth]{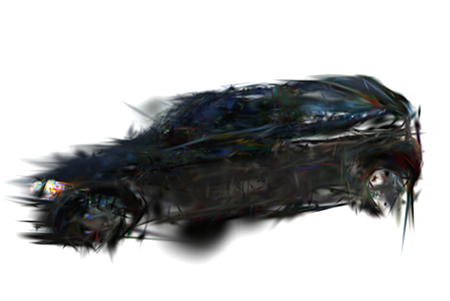} &  
        \includegraphics[width=0.1\textwidth]{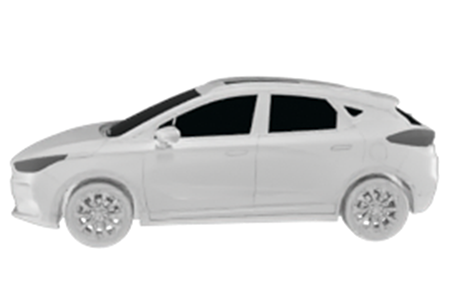} &  
        \includegraphics[width=0.1\textwidth]{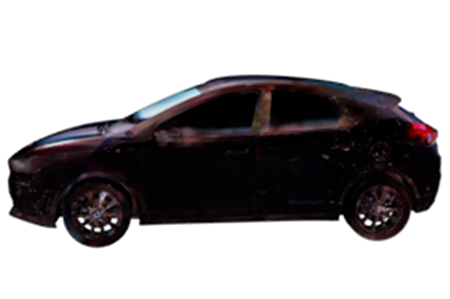} &  
        \includegraphics[width=0.1\textwidth]{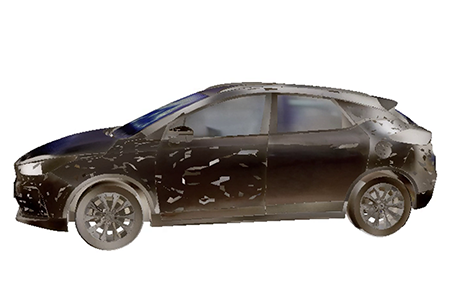} &  
        \includegraphics[width=0.1\textwidth]{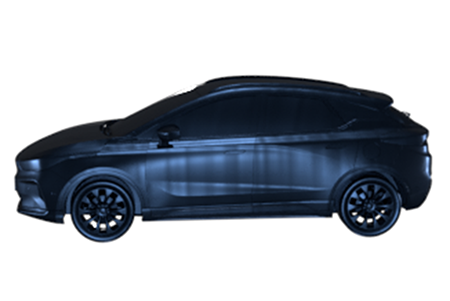} &  
        \includegraphics[width=0.1\textwidth]{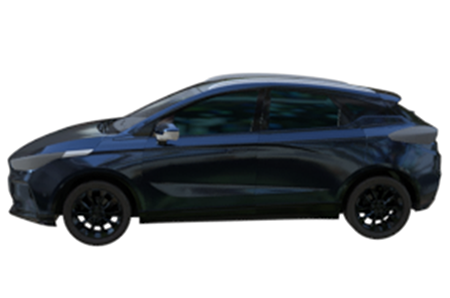}\\ 
    \rotatebox[origin=c]{90}{\small{Ref.}}    &
        \includegraphics[width=0.1\textwidth]{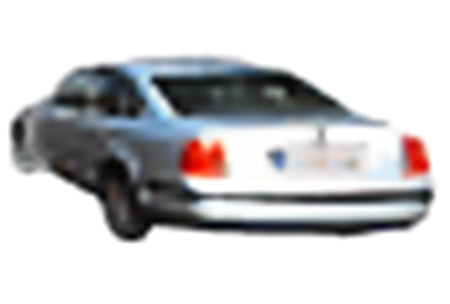} &  
        \includegraphics[width=0.1\textwidth]{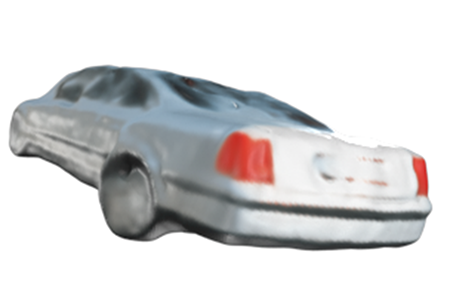} &  
        \includegraphics[width=0.1\textwidth]{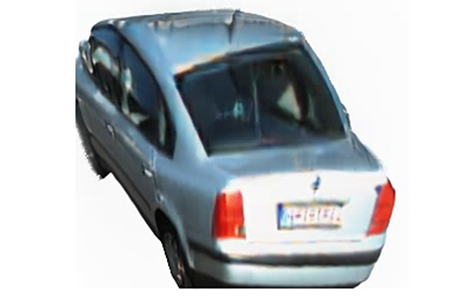} &  
        \includegraphics[width=0.1\textwidth]{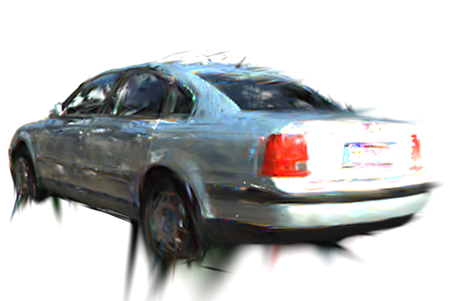} &  
        \includegraphics[width=0.1\textwidth]{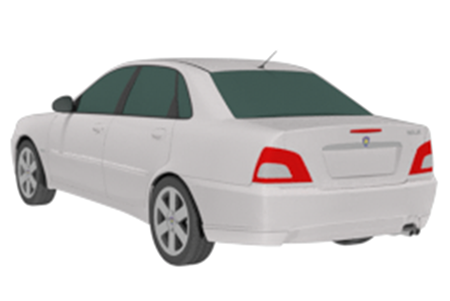} &  
        \includegraphics[width=0.1\textwidth]{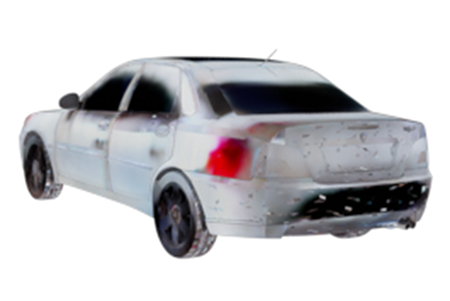} &  
        \includegraphics[width=0.1\textwidth]{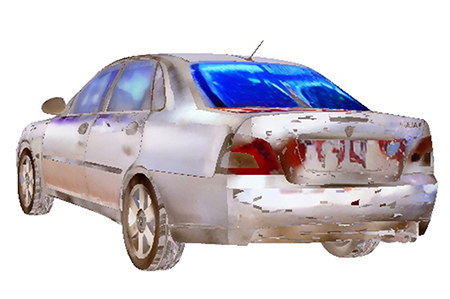} &  
        \includegraphics[width=0.1\textwidth]{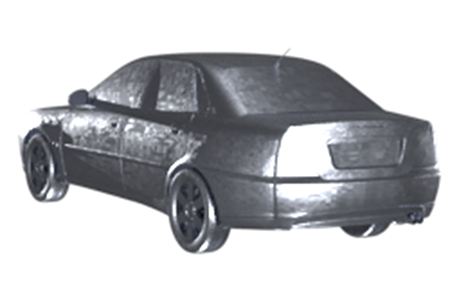} &  
        \includegraphics[width=0.1\textwidth]{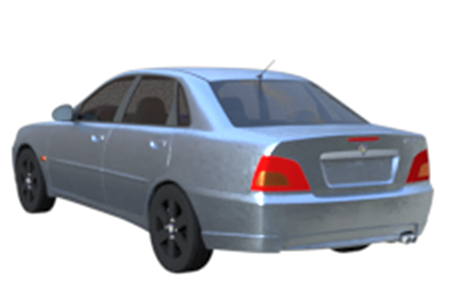}\\ 
    \rotatebox[origin=c]{90}{\small{Rot.}}    &
        \includegraphics[width=0.1\textwidth]{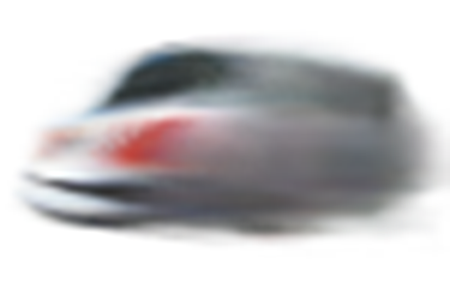} &  
        \includegraphics[width=0.1\textwidth]{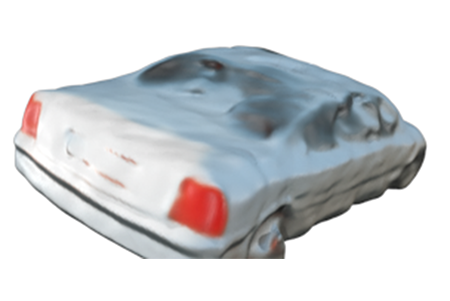} &  
        \includegraphics[width=0.1\textwidth]{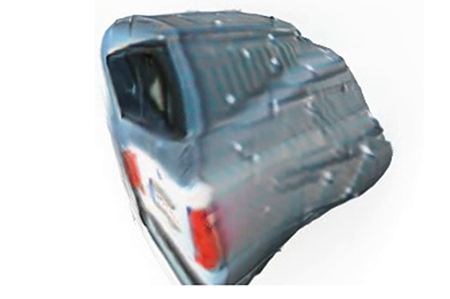} &  
        \includegraphics[width=0.1\textwidth]{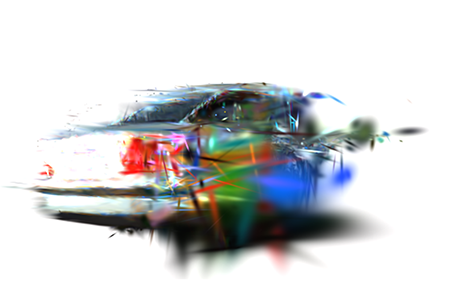} &  
        \includegraphics[width=0.1\textwidth]{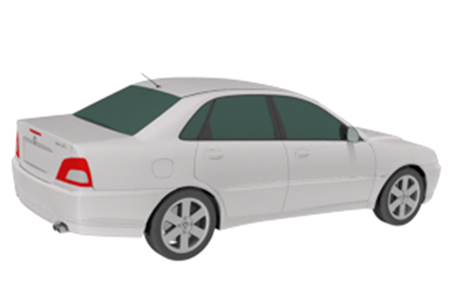} &  
        \includegraphics[width=0.1\textwidth]{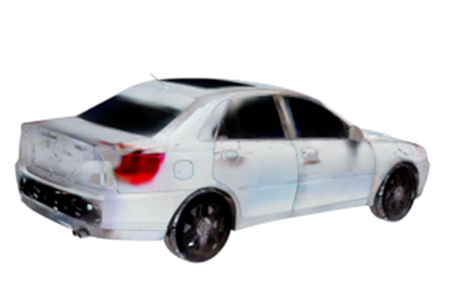} &  
        \includegraphics[width=0.1\textwidth]{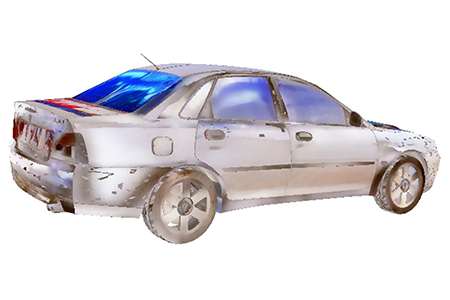} &  
        \includegraphics[width=0.1\textwidth]{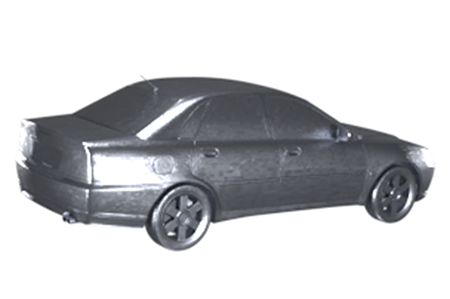} &  
        \includegraphics[width=0.1\textwidth]{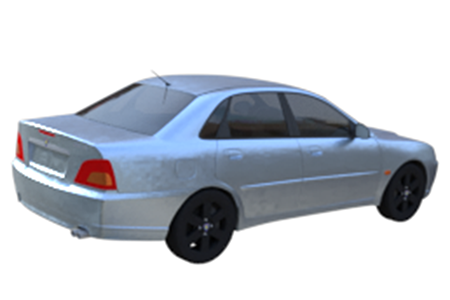}\\
        \vspace{-1ex} 
        & \footnotesize{PixNeRF~\cite{Yu2020pixelNeRFNR}} & \footnotesize{Wonder3D~\cite{Long2023Wonder3DSI}} &
        \footnotesize{LRM~\cite{hong2023lrm}} &
        \footnotesize{HUGS~\cite{zhou2024hugs}} & \footnotesize{Ours (w/o opt.)} & 
        \footnotesize{LatentPaint$^{\dag}$~\cite{Metzer2022LatentNeRFFS}} &
        \footnotesize{Paint3D$^{\dag}$\cite{Zeng2023Paint3DPA}} & \footnotesize{PhotoScene$^{\dag}$\cite{yeh2022photoscene}} & 
        \footnotesize{Ours}\\
    \end{tabular}  
    \caption{\textbf{More qualitative results} on KITTI-360 for novel view synthesis from reference (Ref.) and rotated (rot.) viewpoints.}  
    \label{fig:more_app_comparison}  
\end{figure*}

\begin{figure}[htbp]  
    \centering  
    \includegraphics[width=0.99\linewidth]{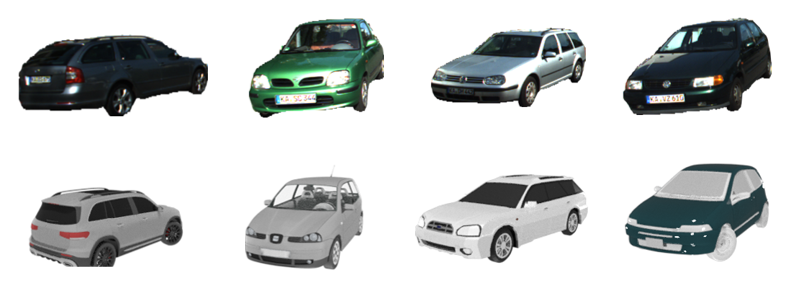}  
    \caption{Pose matching results. It shows that the pose of retrieved CAD models (second row) can match accurately with the pose of the input vehicle images (first row) despite the large difference between appearance and geometry.  }  
    \label{fig:pose_matching}  
\end{figure} 

\subsection{ Pose Matching } \label{sec:pose_matching}

Following~\cite{Chen2024Learning3G,Dai2024ACDCAC}, we choose the CAD model rendering poses based on the DINO~\cite{Amir2021DeepVF} feature similarity with the reference image. First, we crop the vehicles from both the reference image $\bI_{ref}$ and 360$^\circ$ retrieved CAD model renderings  $\left\{\bI_{cad}^k\right\}_{k=1}^M$, where $M = 360/A$ is the number of rendering views, and resize them to the same resolution. Then, we compute the DINO feature maps ~\cite{Amir2021DeepVF} for both vehicle image in reference view and CAD models rendering results using the DINO-ViT encoder $\cE_\text{DINO}$: $\bF_{ref} = \cE_\text{DINO}\left(\bI_{ref}\right)$, $\left\{\b F_{cad}^k\right\}_{k=1}^M = \cE_\text{DINO}(\left\{\b I_{cad}^k\right\}_{k=1}^M)$. Finally, we compute the L2 distances between the $\bF_{ref}$ and the $\left\{\b F_{cad}^k\right\}_{k=1}^M$ and select the rendering that has the minimum L2 distance with the vehicle in the reference view. In our experiment, we find this approach can achieve accurate pose-matching results regardless of appearance and geometry differences between the retrieved CAD models and reference vehicles. The quality results of our pose-matching method are shown in \figref{fig:pose_matching}.

\begin{figure*}[t]  
    \centering  
    \includegraphics[width=0.98\textwidth]{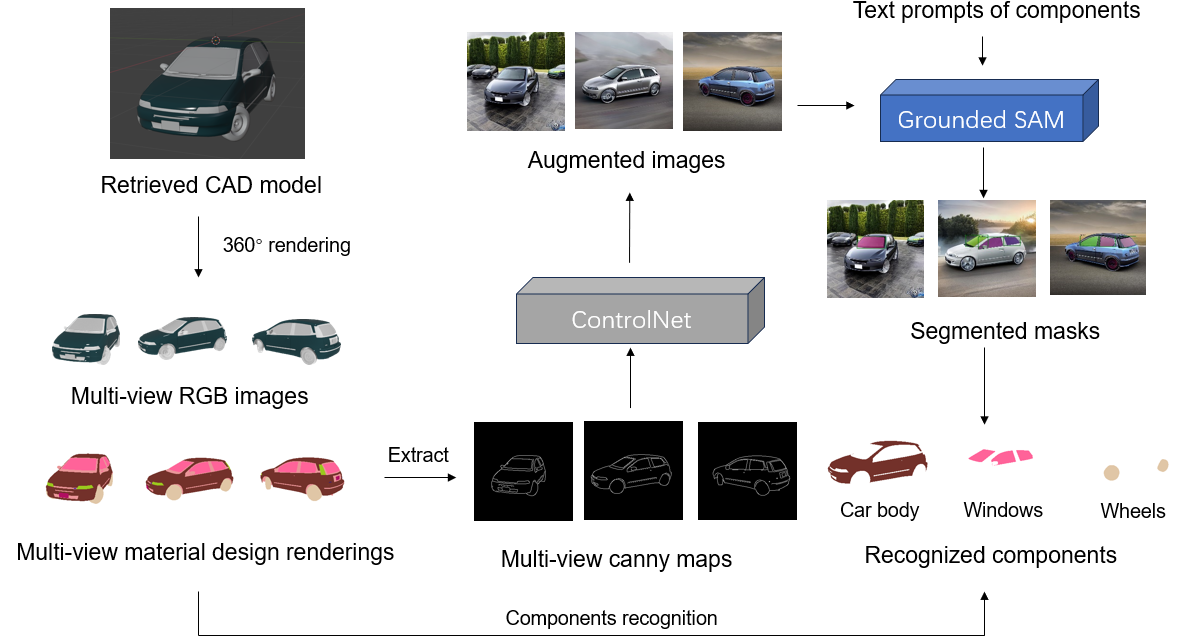}  
    \caption{Illustration of Semantic-based Part-aware Material Prior Retrieval Module. To accurately recognize the semantic meaning of the retrieved CAD model for material prior retrieval, we first render the multi-view material designs and convert them to the canny maps. Subsequently, we use the canny-based ControlNet~\cite{Zhang2023AddingCC} to produce multi-view augmented images. Note that the components' locations in augmented images are aligned with the corresponding material design renderings. After that, we use Grounded SAM~\cite{Ren2024GroundedSA} and components' names (e.g. windows) to segment the components in the augmented images and obtain multi-view segmented masks with corresponding components' meanings. Finally, we utilize these segmented masks to recognize the material indexes of corresponding components in the material designs.}  
    \label{fig:part_ass}  
\end{figure*}

\begin{figure}[htbp]
    \centering
    \includegraphics[width=0.9\linewidth]{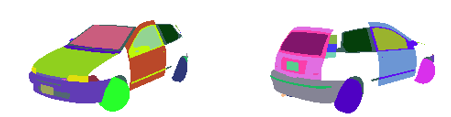}
    \vspace{-1em}
    \caption{\textbf{Disentangled geometry of handcrafted CAD model}. Different colors represent different disentangled geometry.}
    \vspace{-1em}
    \label{fig:dis_geometry}
\end{figure}

\begin{figure}
    \centering
    \includegraphics[width=0.98\linewidth]{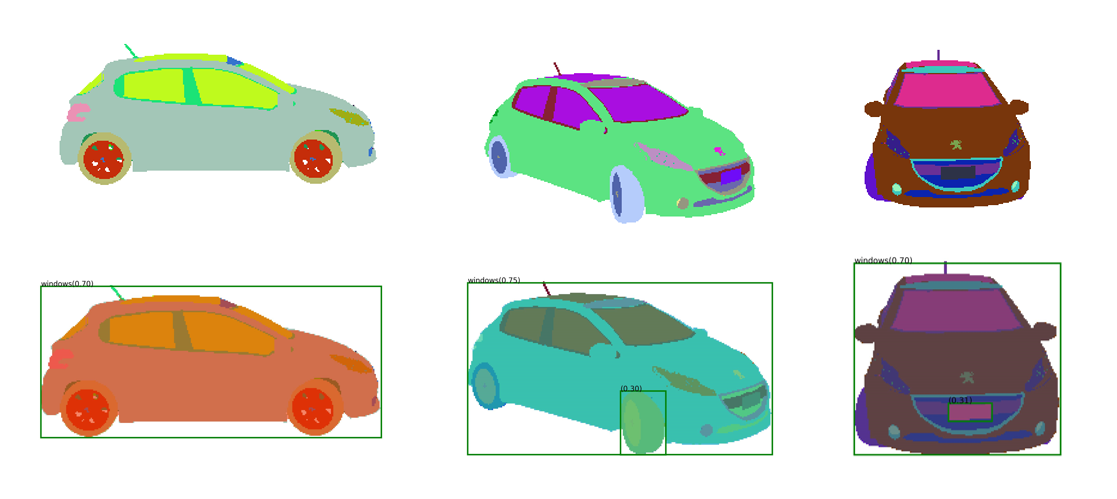}\\
    \includegraphics[width=0.98\linewidth]{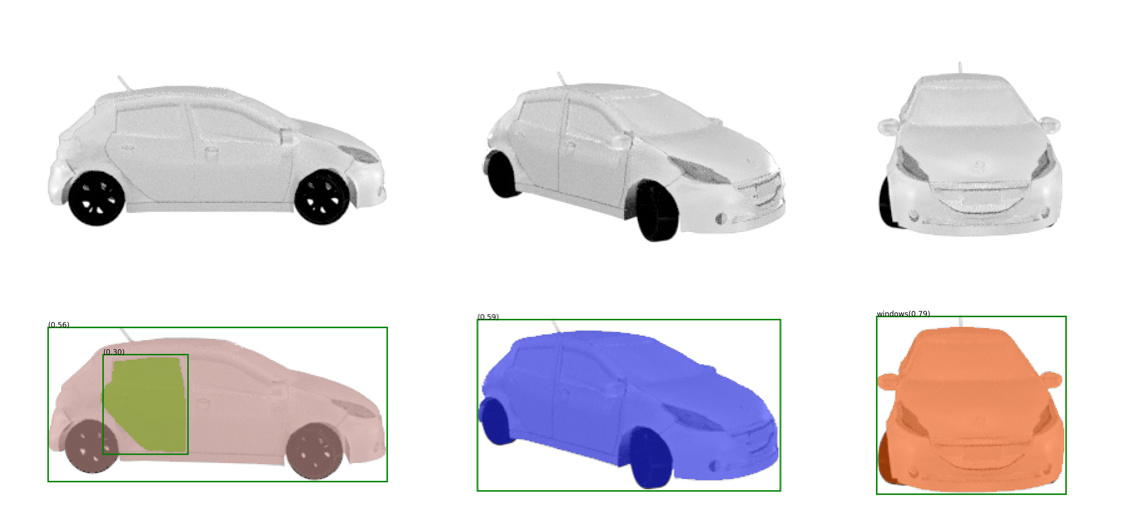}
    \caption{\textbf{Quality results} of part-recognition based on randomly colored material design (top) and retrieved CAD renderings without ControlNet augmentation (bottom) with the text prompt of "windows".}
    \label{fig:groundsam_openshape}
\end{figure}

\subsection{Part-level Material Prior Retrieval} \label{sec:mat_prior_retrieval}

Since the retrieved CAD models usually have an unsatisfactory appearance, simply using Grounded SAM to segment the CAD model renderings will lead to many failure cases. ControlNet can translate primitives like edges into realistic pictures. Therefore, we propose to use ControlNet to augment the CAD model renderings and ensure the accurate segmentation of Grounded SAM. Specifically, we first extract edges from the material index maps rendered in 360$^\circ$. Then, we input edges into a canny-based pre-trained ControlNet model and obtain the augmented multi-view images. Note that this canny-based ControlNet translation does not affect the position of the components. After that, we use Grounded SAM to segment the 360$^\circ$ augmented images with component text prompts like windows and wheels. Once we get the multi-view segmentation results, we first select the rendering with the highest mean mask confidence. Then, we calculate the material index masks that have an intersection with the segmented mask. We define them as active materials $\b{Mat}_{act}$. We calculate the masks of each active material $\b{Mat}_{act}$ in material index map $\b M_{ind}$ and in the Grounded SAM segmentation map $\b M_{seg}$. We then compute the IOU between $\b M_{ind}$ and $\b M_{seg}$. If the IOU is larger than the IOU threshold (we set the IOU threshold as 0.5), the material will be classified into the corresponding component. We illustrate our method in \figref{fig:part_ass}.

\subsection{Material Design Merging Using DINO Feature} \label{sec:car_body_merging}

Since the bodies of some vehicles are composed of many small components in the CAD models, only retrieving and optimizing materials for the largest part will lead to unsatisfactory results. However, Grounded SAM sometimes can't recognize tiny components like vehicle lights. Simply regarding all remaining parts after component recognition as car bodies will also lead to inaccurate material assignment.  To this end, we utilize the DINO corresponding points proposed in~\cite{Amir2021DeepVF} to merge the small components in the CAD models. Specifically, we first segment the known components in the input image using Grounded SAM. Then, we calculate the corresponding points between the remaining parts in the input images and the CAD model renderings. Since the remaining parts in the input images are the car body, the corresponding parts in the CAD model renderings are the car body as well. Besides, with a suitable setting of corresponding points' numbers, tiny components not belonging to car bodies will not be wrongly merged. During our experiment, this kind of merging produces good material assignment results on tiny components of CAD models.

\subsection{Material Optimization} \label{sec:mat_optimization_appendice}
Since there is no exact correspondence between rendered and reference pixels, we use a part-level loss $\b \ell_{stat}$ by minimizing the difference between the mean and variance of the corresponding parts following ~\cite{yeh2022photoscene}:

\begin{align}
\b \ell_{mean} & = | \mu(\b I_{ref} \cdot \b S_{ref}[\b c]) - \mu(\widehat{\b I}_{render} \cdot \b S_{cad}[\b c]) | \\
\b \ell_{var} & = | \sigma ^ 2(\b I_{ref} \cdot \b S_{ref}[\b c]) - \sigma ^ 2(\widehat{\b I}_{render} \cdot \b S_{cad}[\b c]) | \\
\b \ell_{stat} & = \b \ell_{mean} + \b \ell_{var}
\end{align}

where $\widehat{\b I}_{render}$ is the CAD model rendering after pose matching, $\b S_{ref}[\b c]$ and $\b S_{cad}[\b c]$ are the segmentation masks of the component $\b c$ in the reference view and CAD model rendering.

To match the patterns of the reference view, we use a masked VGG loss $\ell_{VGG}$ using Gram matrices \cite{Gatys2016ImageST} to enhance visual similarity:

\begin{equation}
\b \ell_{vgg} = | Gram(\b I_{ref}, \b S_{ref}[\b c]) - Gram(\widehat{\b I}_{render}, \b S_{cad}[\b c]) |
\end{equation}

To further match the color of the reference vehicles, we add a masked RGB loss $\ell_{rgb}$ on the overlap region between components in the reference view and CAD model rendering:

\begin{equation}
\b \ell_{rgb} = \left| \mathbf{I}_{ref} \cdot \mathbf{S}_{overlap} - \mathbf{I}_{cad} \cdot \mathbf{S}_{overlap} \right|
\end{equation}
The total loss function is shown as below:

\begin{equation}
\b{\ell_{total}} = \b{\lambda_{stat}} \b{\ell_{stat}} + \b{\lambda_{vgg}} \b{\ell_{vgg}} + \b{\lambda_{rgb}} \b{\ell_{rgb}}
\end{equation}

\begin{figure}
    \centering
    \includegraphics[width=0.9\linewidth]{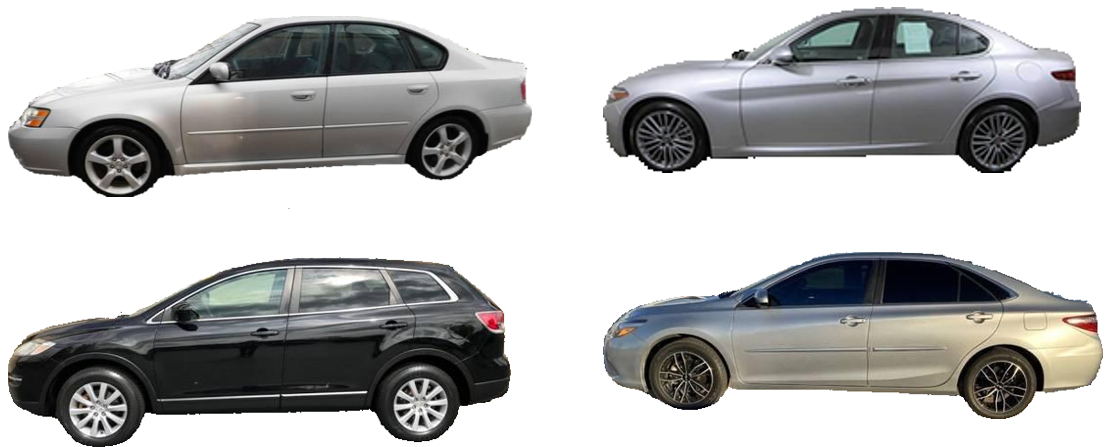}
    \caption{Two types of car body materials with different roughness. Vehicles in the left column are painted and vehicles in the right column are not painted.}
    \label{fig:diff_rough}
    \vspace{-1em}
\end{figure}

In our experiment, we set the $\lambda_{stat}$ to 0.1, the $\lambda_{vgg}$ to 1, the $\lambda_{rgb}$ to 1. Note that spatially varying roughness parameters are difficult to optimize from single-view images due to limited highlight observations. Handcrafted procedural material graphs provide photorealistic spatially varying effects, so the roughness parameters of the retrieved material prior are fixed during optimization, as in \cite{yeh2022photoscene}. Besides, we observe two types of materials with distinct spatially varying effects in car bodies depending on whether the vehicles are painted or not, as shown in \figref{fig:diff_rough}. To best fit the observation, we recommend selecting the corresponding car body material prior via the user interface.

\begin{figure*}[htbp]
    \centering
    \includegraphics[width=0.95\textwidth]{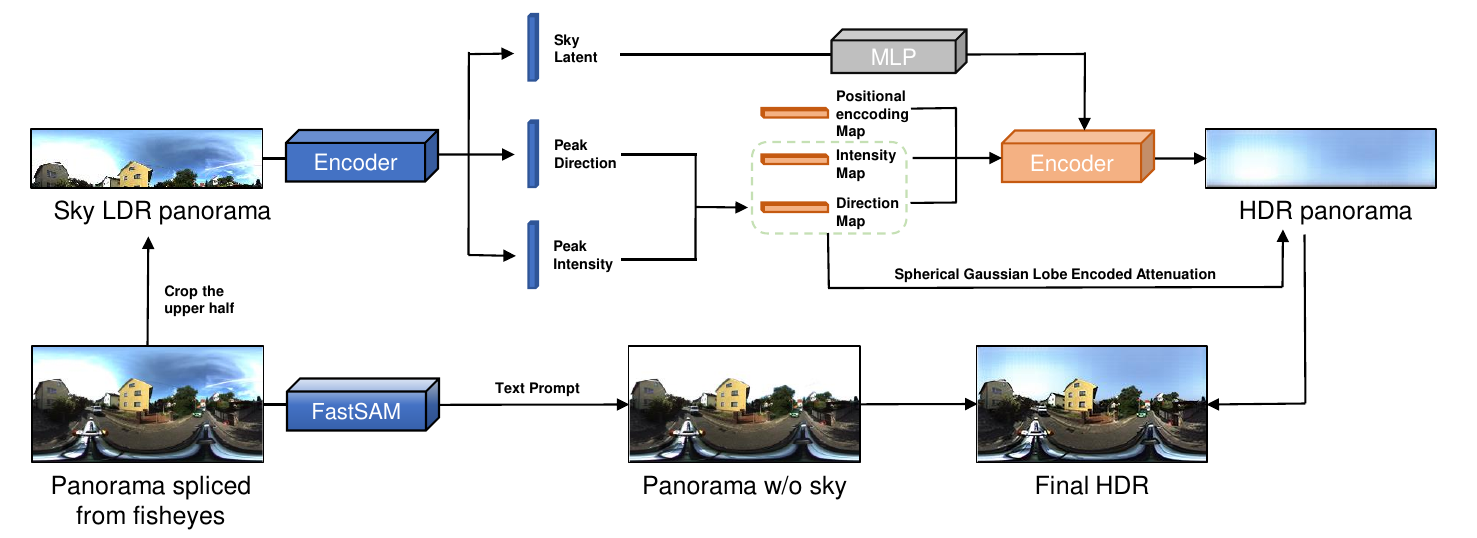}
    \caption{LDR to HDR reconstruction pipeline. The upper half obtains the HDR panorama from the LDR input. The other half stitches the origin panorama with the predicted HDR sky to get the spatially varying lighting}
    \label{fig:light_imp}
\end{figure*}

\subsection{ Spatially Varying Lighting Estimation Based on Fisheye Images} \label{sec:lighting}
As shown in \figref{fig:light_imp}, to obtain spatially varying lighting, we first stitch 2 fisheye images into an LDR panorama. Then we crop the upper part of the panorama representing the skydome and feed it into the ChatSim ~\cite{Wei2024EditableSS} LDR to HDR prediction network. After obtaining the HDR panorama of the sky part, we use FastSAM ~\cite{Zhao2023FastSA} with text prompts to obtain the ground part. FastSAM selectively ignores detailed pixels, enabling us to separate the clean sky, which could be beneficial to subsequent usage. After performing numerical correction on the LDR image and concatenating it with the previously obtained HDR panorama, we can obtain the local lighting of the current position where the fisheye image is captured.

\subsection{ Background Reconstruction using 3DGS} \label{sec:background_rec}
We employ the HUGS \cite{zhou2024hugs} to reconstruct the background of urban scenes. This process involves utilizing multi-view appearance observations and pseudo-semantic labels obtained from InverseForm \cite{borse2021inverseform}. The HUGS is trained for a total of 30,000 iterations, using two front-perspective cameras and two side-look fisheye cameras in each sequence. Each sequence encompasses 40 frames both prior to and following the target frame. For this reconstruction process, we adhere to the configurations defined by the HUGS. Notably, when converting a static car into our optimized CAD model, we can use inpainting methods~\cite{Yu2023InpaintAS} for background inpainting to animate the car without leaving holes in the ground.

\section{ Implementation Details of Experiments} \label{sec:exp_detail}

\subsection{ Baselines Implementation} \label{sec:base_imp}

During appearance comparison, we evaluate 1800 images of 30 models rendered from 360$^\circ$ views and report FID/KID scores comparing with 1800 reference images collected from  ~\cite{Yang2015ALC}. Besides, we report the LIPIS scores by comparing the difference between input reference vehicle images and CAD renderings under matched poses.

\noindent \textbf{PixelNeRF. } PixelNeRF ~\cite{Yu2020pixelNeRFNR} is an image-based reconstruction method using a conditional implicit function. It supports single-view reconstruction tasks on real-world images. We use the official model pre-trained on ShapeNet ~\cite{Chang2015ShapeNetAI} to evaluate the performance. We input our single-view images to the PixelNeRF and rendered the reconstructed neural radiance field in 360$^\circ$ with 180 frames.

\noindent \textbf{Wonder3D. } Wonder3D ~\cite{Long2023Wonder3DSI} is a image-based single view 3D generation method using diffusion priors. We use the official pretrained model to evaluate its single-view generation quality on our input images.

\noindent \textbf{LRM. } LRM~\cite{hong2023lrm} is a conditional implicit function based single view 3D reconstruction method with large scale training. Since the official LRM implementation hasn't been open-sourced, we use the open-sourced implementation OpenLRM~\cite{openlrm}. When inferring on single view image, we simply use its open-sourced pre-trained model. 

\noindent \textbf{HUGS. }
As described in \ref{sec:background_rec}, we employ HUGS to reconstruct the urban scene, including the target vehicle. The extraction of the target vehicle requires identifying the specific Gaussians that constitute the vehicle. Fortunately, our approach achieved the 3D semantic reconstruction facilitated by HUGS, where every 3D Gaussian possesses a semantic label. This allows for extracting the target vehicle by selecting 3D Gaussians that lie within the bounding box and carry car semantic labels. By manipulating the position and orientation of the 3D Gaussians with a transformation matrix, we can easily manipulate the vehicle representation.

\begin{table}[t]
\centering
\resizebox{.49\textwidth}{!} 
{
\begin{tabular}{c|c|ccc}
\toprule
 Labor Cost & Method & FID$\downarrow$ & KID$\downarrow$ & LPIPS$\downarrow$\\
\midrule
 \text{High} & OpenShape~\cite{liu2024openshape} & 73.10 & \textbf{0.0453} & 0.5761\\
 \text{Middle} & UrbanCAD (w/o opt.) & 81.05 & 0.0567 & 0.6174\\
 \text{Low} & OpenShape$^{\star}$~\cite{liu2024openshape} & 116.36 & 0.0990 & 0.6676\\
 \text{Middle} & UrbanCAD (Ours) & \textbf{62.80} & 0.0479 & \textbf{0.5242}\\
  \bottomrule
\end{tabular}
}
\caption{\textbf{Quantitative Comparison} on the photorealism of retrieved CAD models with different kinds of materials. }
\vspace{-1em}
\label{tab:app_supple}

\end{table}

\noindent \textbf{UrbanCAD (w/o opt.). } UrbanCAD (w/o opt.) is implemented by directly using the official pre-trained checkpoint of Openshape ~\cite{liu2024openshape}, a multi-modality joint representation method, to retrieve the CAD models from Objaverse ~\cite{deitke2023objaverse} dataset according to the input single-view images. Note that while Objaverse includes vehicle CAD models with high-quality texture maps, these require significant manual labor and cannot be optimized to fit observation data. In contrast, our method only requires CAD models with base colors as input, reducing the need for human effort. We further evaluate the quality of these labor-intensive handcrafted textures in \tabref{tab:app_supple}. OpenShape \cite{liu2024openshape} refers to the retrieved CAD models with external handcrafted texture maps, while UrbanCAD (w/o opt.) refers to the CAD models with base colors. OpenShape$^{\star}$ \cite{liu2024openshape} denotes the retrieved CAD models without any materials. UrbanCAD (Ours) refers to CAD models with our optimized materials. The results show that our method generates materials that better fit the observations, achieving comparable or superior quality to the labor-intensive handcrafted texture maps.

\noindent \textbf{LatentPaint. } LatentPaint~\cite{Metzer2022LatentNeRFFS} is a mesh texturing method using a generative model. When implementing LatentPaint, we found its open-sourced code doesn't support textual inversion. Therefore, we use ChatGPT4~\cite{Achiam2023GPT4TR} to implement textual inversion by asking ChatGPT4 to estimate the colors of the input vehicles. After we get the colors described in the text, we use the official implementation of LatentPaint to accomplish the mesh texturing task.

\noindent \textbf{Paint3D. } Paint3D~\cite{Zeng2023Paint3DPA} is a SOTA mesh texturing method using diffusion model. It generates high-resolution textures in a coarse-to-fine manner and supports texutre transfer from a single view image using IP-Adapter~\cite{Ye2023IPAdapterTC}. In our implementation, we directly use its open-source code and checkpoints to do the inference.

\noindent \textbf{PhotoScene. } Since the procedural graph library used in PhotoScene is different from our method, which may lead to unfairness, we implement PhotoScene on our pre-defined procedural graph library. Specifically, we directly assign the metal material used in our method and further optimize the material, since retrieving materials based on visual similarity proposed in Photoscene will lead to severe degradation of appearance.

\begin{figure*}[htbp]  
    \centering 
    \begin{minipage}[b]{0.98\textwidth}
        \begin{minipage}[b]{0.32\textwidth}
            \centering
            \includegraphics[width=1\textwidth]{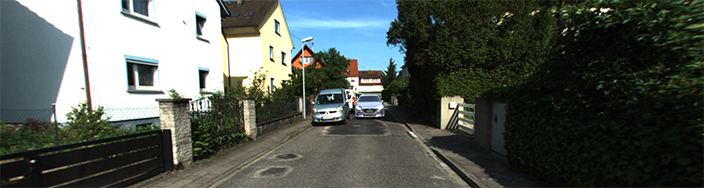}
        \end{minipage}
        \hfill
        \begin{minipage}[b]{0.32\textwidth}
            \centering
            \includegraphics[width=1\textwidth]{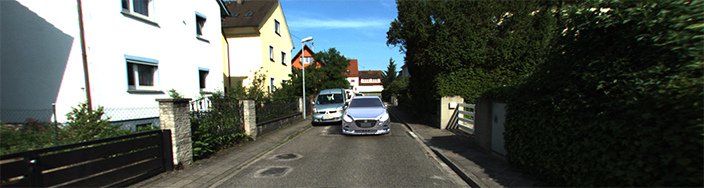}
        \end{minipage}
        \hfill
        \begin{minipage}[b]{0.32\textwidth}
            \centering
            \includegraphics[width=1\textwidth]{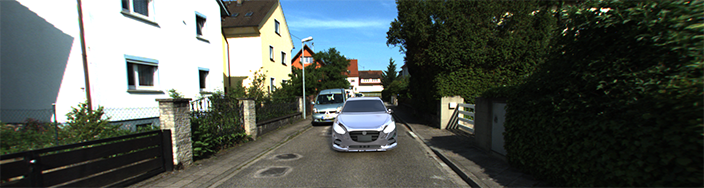}
        \end{minipage}
        \hfill
    \centering 
    \text{Trajector 1}
    \end{minipage}
    \begin{minipage}[b]{0.98\textwidth}
        \begin{minipage}[b]{0.32\textwidth}
            \centering
            \includegraphics[width=1\textwidth]{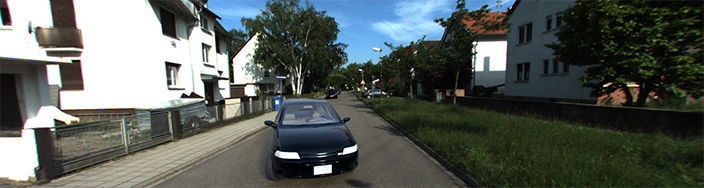}
        \end{minipage}
        \hfill
        \begin{minipage}[b]{0.32\textwidth}
            \centering
            \includegraphics[width=1\textwidth]{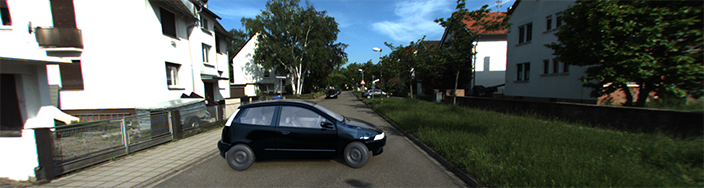}
        \end{minipage}
        \hfill
        \begin{minipage}[b]{0.32\textwidth}
            \centering
            \includegraphics[width=1\textwidth]{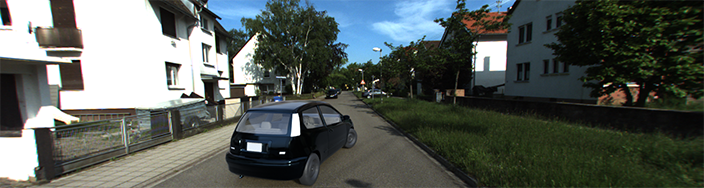}
        \end{minipage}
        \hfill
    \centering
    \text{Trajectory 2}
    \end{minipage}
    \begin{minipage}[b]{0.98\textwidth}
        \begin{minipage}[b]{0.32\textwidth}
            \centering
            \includegraphics[width=1\textwidth]{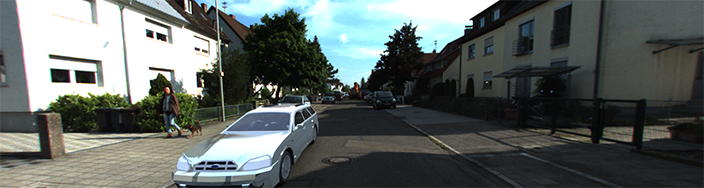}
        \end{minipage}
        \hfill
        \begin{minipage}[b]{0.32\textwidth}
            \centering
            \includegraphics[width=1\textwidth]{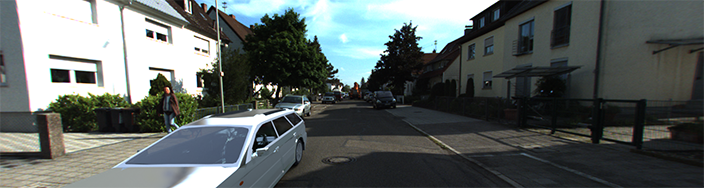}
        \end{minipage}
        \hfill
        \begin{minipage}[b]{0.32\textwidth}
            \centering
            \includegraphics[width=1\textwidth]{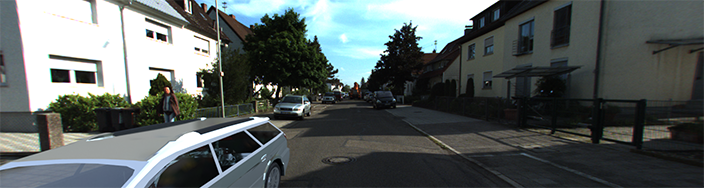}
        \end{minipage}
        \hfill
    \centering 
    \text{Trajectory 3}
    \caption{Illustration of our synthetic data during self-driving system testing.}
    \label{fig:syn_data}
    \end{minipage}
\end{figure*}

\subsection{ Synthetic Data Generation } \label{sec:syn_data}  We utilize a series of 3d bounding boxes to control the movement of vehicles. We construct our synthetic data for self-driving perception system testing in 3 different trajectories as illustrated in \figref{fig:syn_data}. Specifically, trajectory 1 involves vehicles moving normally on the road. Trajectory 2 includes scenarios of vehicles rotating 360$^\circ$. Trajectory 3 involves vehicles moving in near and partially obscured views, which are typically more challenging for perception models. For each group of synthetic data, there are 60 images for Trajectory 1, 90 images for Trajectory 2, and 120 images for Trajectory 3. When constructing scenarios using UrbanCAD without lighting estimation, we position six uniform point lights along the positive and negative x, y, and z axes.

\subsection{ Perception Systems Implementation} \label{sec:per_imp} For YOLOv8 instance segmentation method, we use the official model yolov8n pre-trained on COCO dataset ~\cite{Lin2014MicrosoftCC}. For the Mask2Former instance segmentation method, we use the official pretrianed models with different backbones on the cityscapes dataset ~\cite{Cordts2016TheCD}.

\subsection{ Computing Resource} \label{sec:compute_resource}

We use a single RTX3090 GPU to perform material optimization. Optimizing a material takes about 35 seconds for 300 optimization epochs.

\section{ Functionality } \label{sec:func}

Since our created vehicle models are fully controllable, we showcase more editing results including component editing, relighting, material transfer, 360$^\circ$ rotation, and novel view rendering.

\subsection{Component Editing}

\begin{figure}[ht]  
    \def \mywidth {0.48\textwidth}
    \centering  
    \begin{subfigure}[b]{\mywidth}  
        \centering  
        \includegraphics[width=\textwidth]{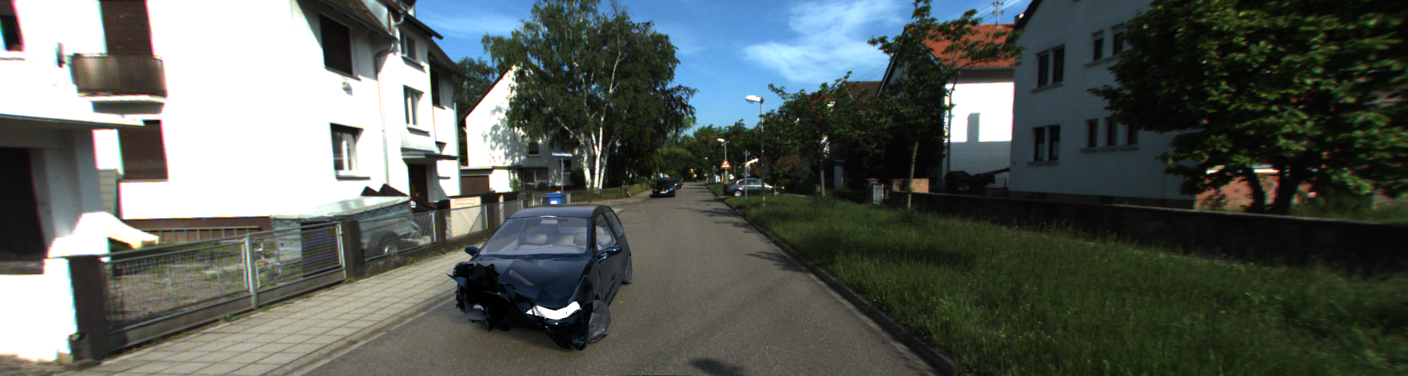}  
        \caption{Car collision.}  
    \end{subfigure}  
    \begin{subfigure}[b]{\mywidth}  
        \centering  
        \includegraphics[width=\textwidth]{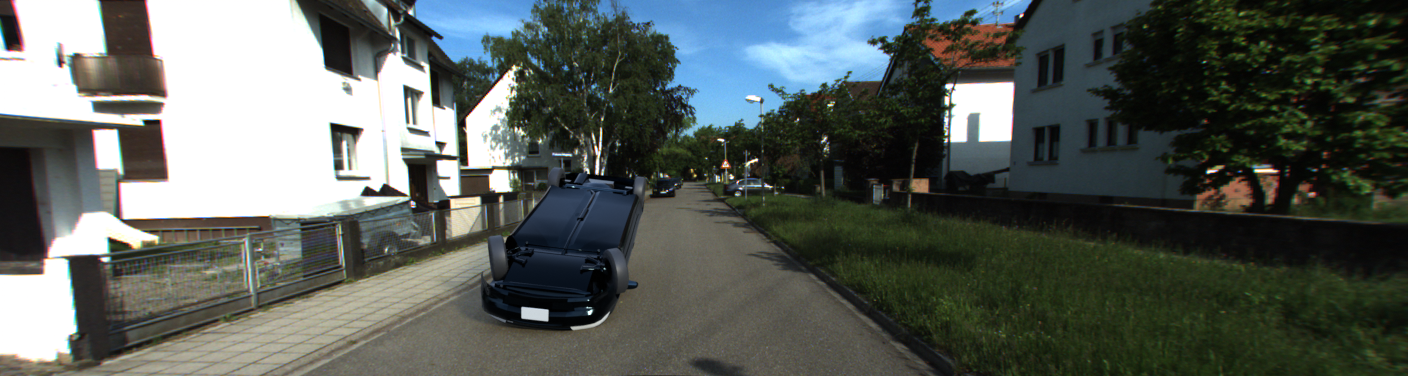}  
        \caption{Car turns upside down.}  
    \end{subfigure} 
    \begin{subfigure}[b]{\mywidth}  
        \centering  
        \includegraphics[width=\textwidth]{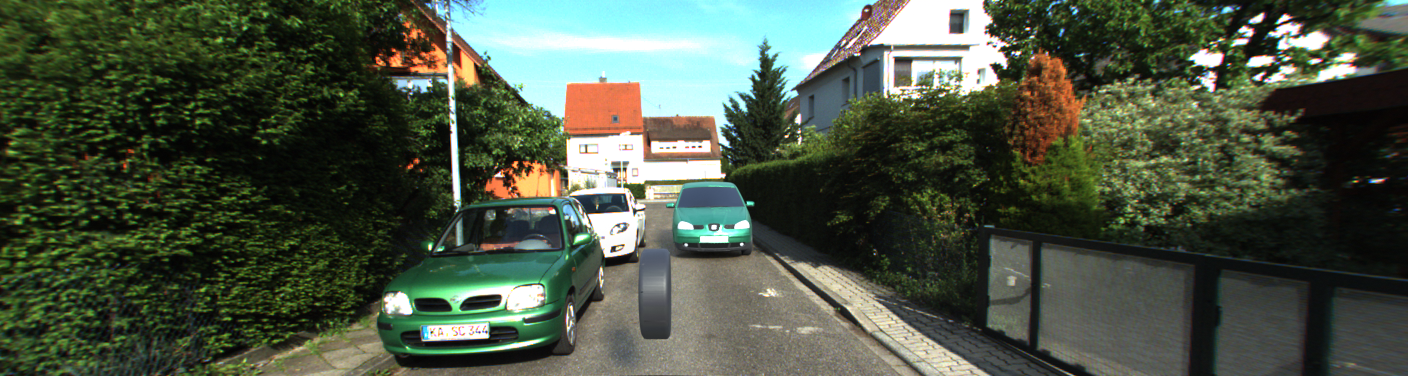}  
        \caption{Tire rolling.}  
    \end{subfigure} 
    \caption{More corner cases.}
    \label{fig:components editing}
\end{figure} 

 Our produced 3D vehicle models support easy component editing mainly due to the handcrafted disentangled geometry as shown in \figref{fig:dis_geometry}. Note that complete component editing requires human effort for animation, such as setting joint types and parameters in Blender. Additionally, some retrieved handcrafted CAD models may have merged geometry, for example, the four wheels are merged in one mesh. For these cases, simply hiding other vehicle components and entering the edit mode to separate the wheels by selection in the Blender can solve the problem with small manual efforts. However, we notice that some vehicle CAD models have been post-processed by geometry merging, which means the loss of part controllability. Fortunately, most handcrafted vehicle CAD models in the Objaverse still preserve part controllability without being post-processed, and many post-processed CAD models still have disconnected geometry, which can be manually separated by Blender ``Separate Selection" operation after selecting connected geometry (``Select Linked" function in ``Select" menu). Besides, more corner case results are displayed in Figure \ref{fig:components editing} thanks to the representation of CAD models. In addition to the editing results mentioned earlier, we can generate more scenes, using the powerful physical simulation effects in Blender. By assigning physics properties to the vehicle model, we can create collision scenes or even simulate car accidents in Blender.

\subsection{Relighting}

\begin{figure}[h]
\centering
\includegraphics[width=0.48\textwidth]{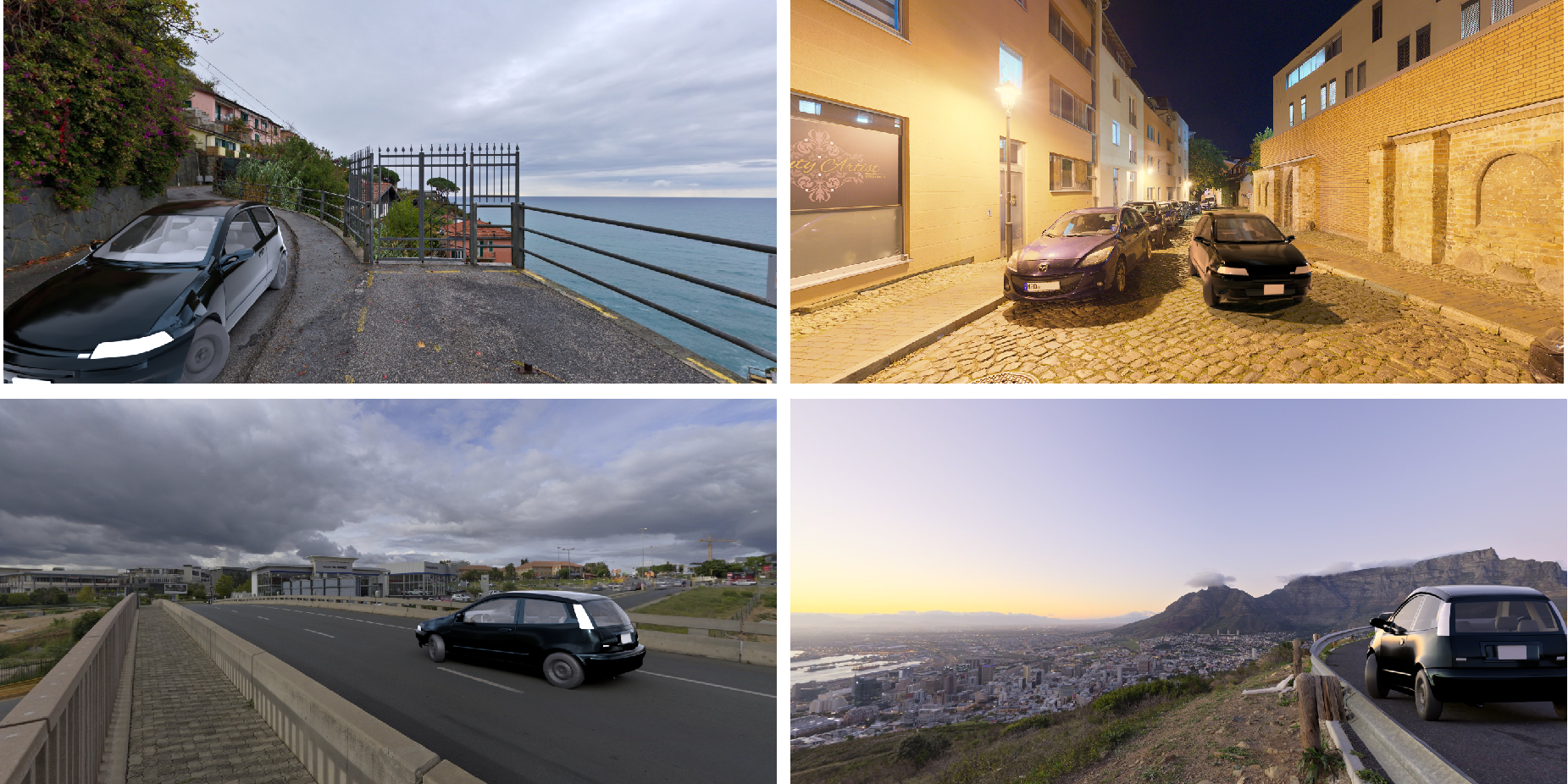}
\caption{ Realistic Insertion.}
\label{fig:realisic insertion}
\end{figure}

Realistic insertion results are shown in Figure \ref{fig:realisic insertion}. We utilize the LDR and HDR pairs from online databases to perform the relighting.

\subsection{Material transfer}

\begin{figure}[ht]  
    \centering  
    \begin{subfigure}[b]{0.48\textwidth}  
        \centering  
        \includegraphics[width=\textwidth]{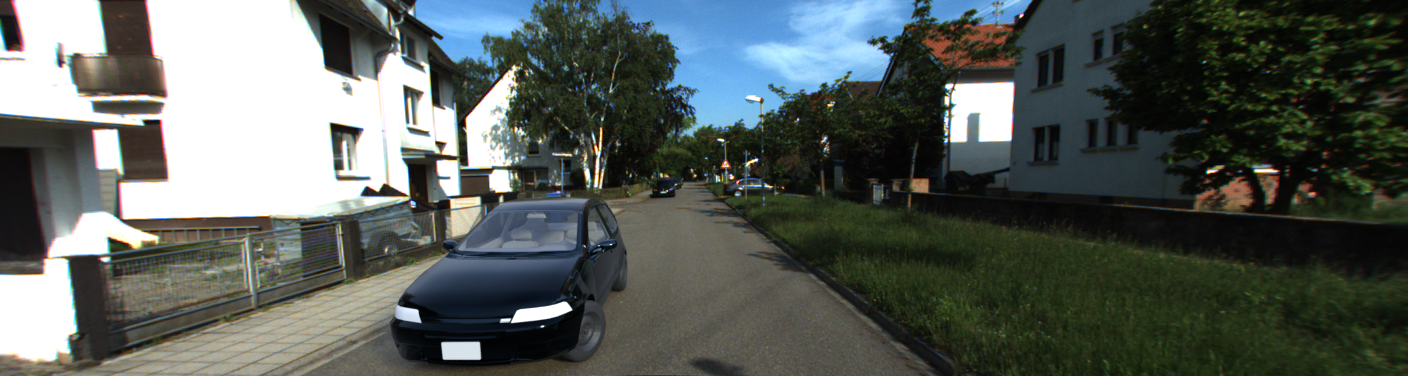}  
        \caption{Target material}  
    \end{subfigure}  
    \begin{subfigure}[b]{0.48\textwidth}  
        \centering  
        \includegraphics[width=\textwidth]{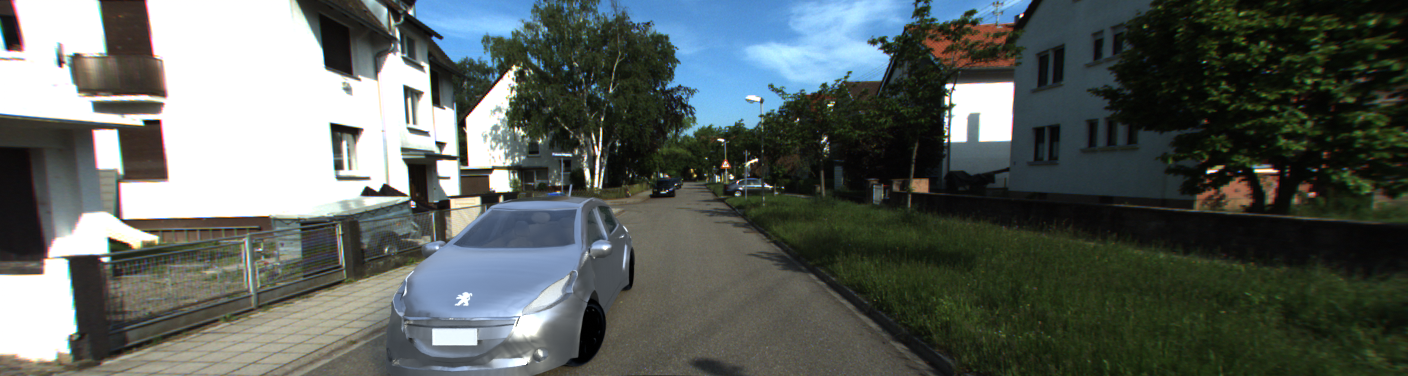}  
        \caption{Source model}  
    \end{subfigure} 
    \begin{subfigure}[b]{0.48\textwidth}  
        \centering  
        \includegraphics[width=\textwidth]{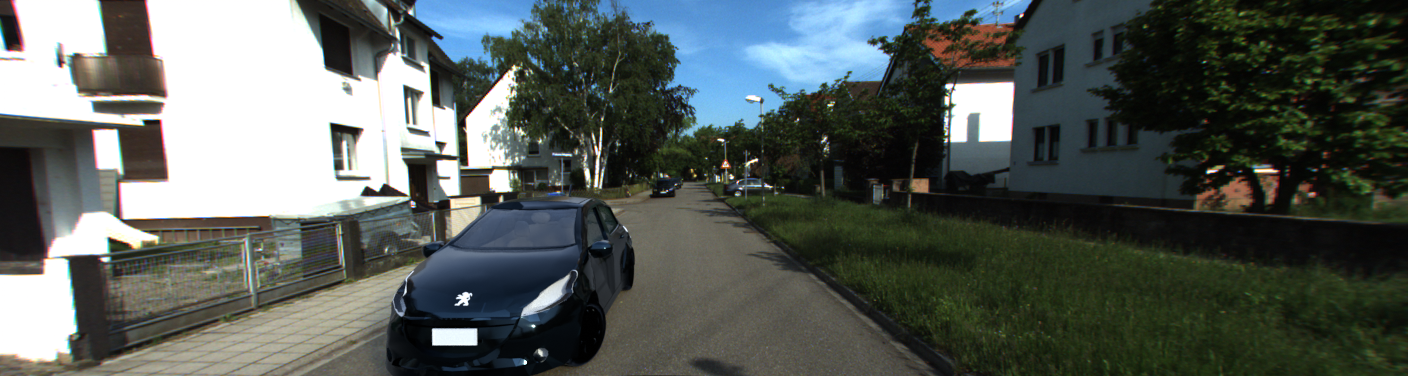}  
        \caption{Material transferred model}  
    \end{subfigure} 
    \caption{Material Transfer.}
    \label{fig:material transfer}
\end{figure} 

Material transfer results are shown in Figure \ref{fig:material transfer}. Since we have obtained the semantic meaning of CAD model material designs, we can easily transfer the part-aware material from one to another.

\begin{figure}[htbp]
    \centering
    \includegraphics[width=0.48\textwidth]{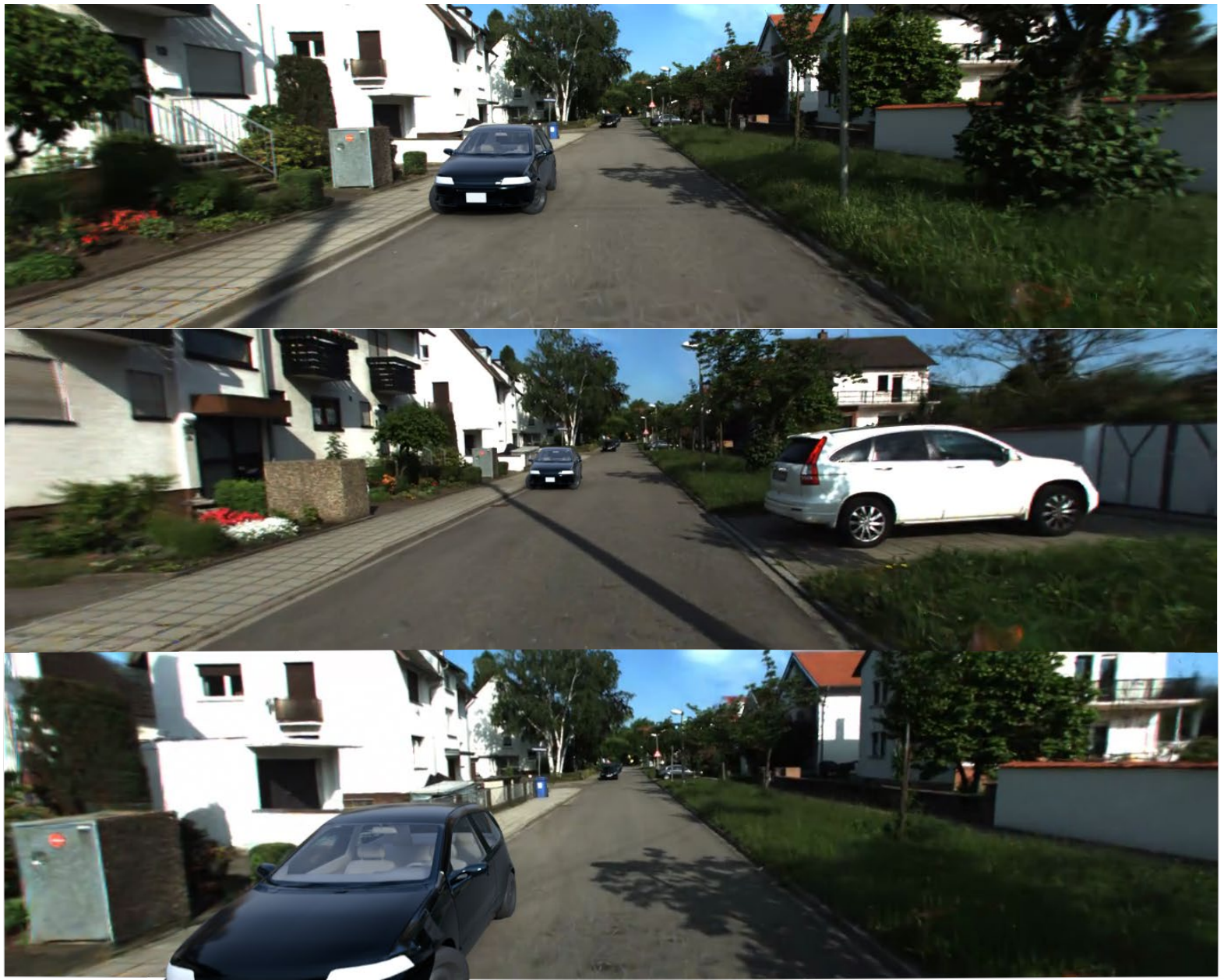}
    \caption{Novel View Synthesis}
    \label{fig:rot}
\end{figure}

\subsection{Novel view rendering}
We showcase our novel view rendering results after reconstructing the background using the implicit function and inserting our produced vehicle model, as shown in \figref{fig:rot}. Our method can produce high-fidelity rendering results of both background scenes and foreground vehicles under novel viewpoints.

\section{ Additional Experiments and Analysis } \label{sec:more}

\subsection{Lighting Estimation Comparison} \label{sec:lighting_comp}

\begin{figure}[htbp] 
    \def \mywidth {0.48\textwidth}
    \centering  
    \begin{subfigure}[b]{\mywidth}  
        \centering  
        \includegraphics[width=0.95\textwidth]{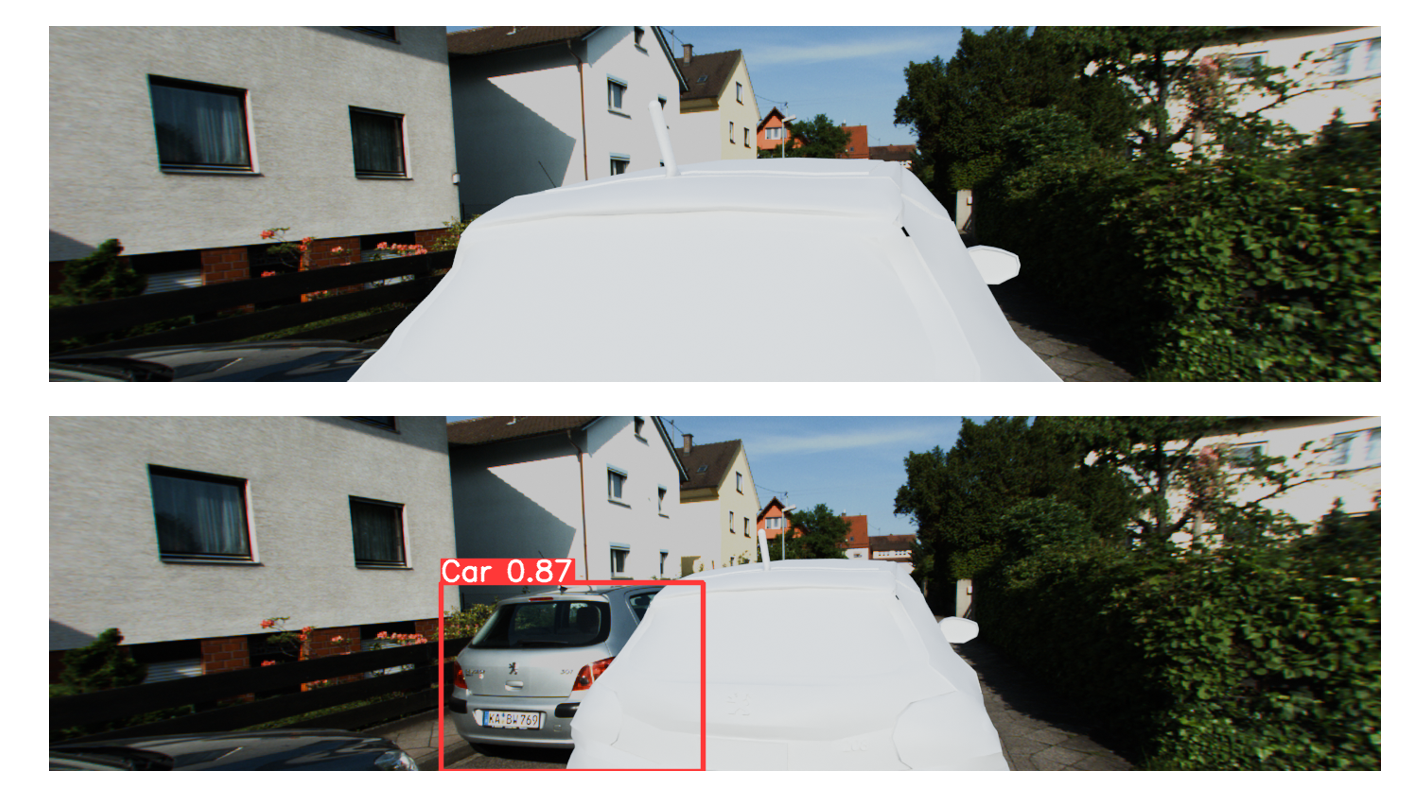}  
        \caption{Scenarios with CAD model without opitmization}  
    \end{subfigure}  
    \begin{subfigure}[b]{\mywidth}  
        \centering  
        \includegraphics[width=0.95\textwidth]{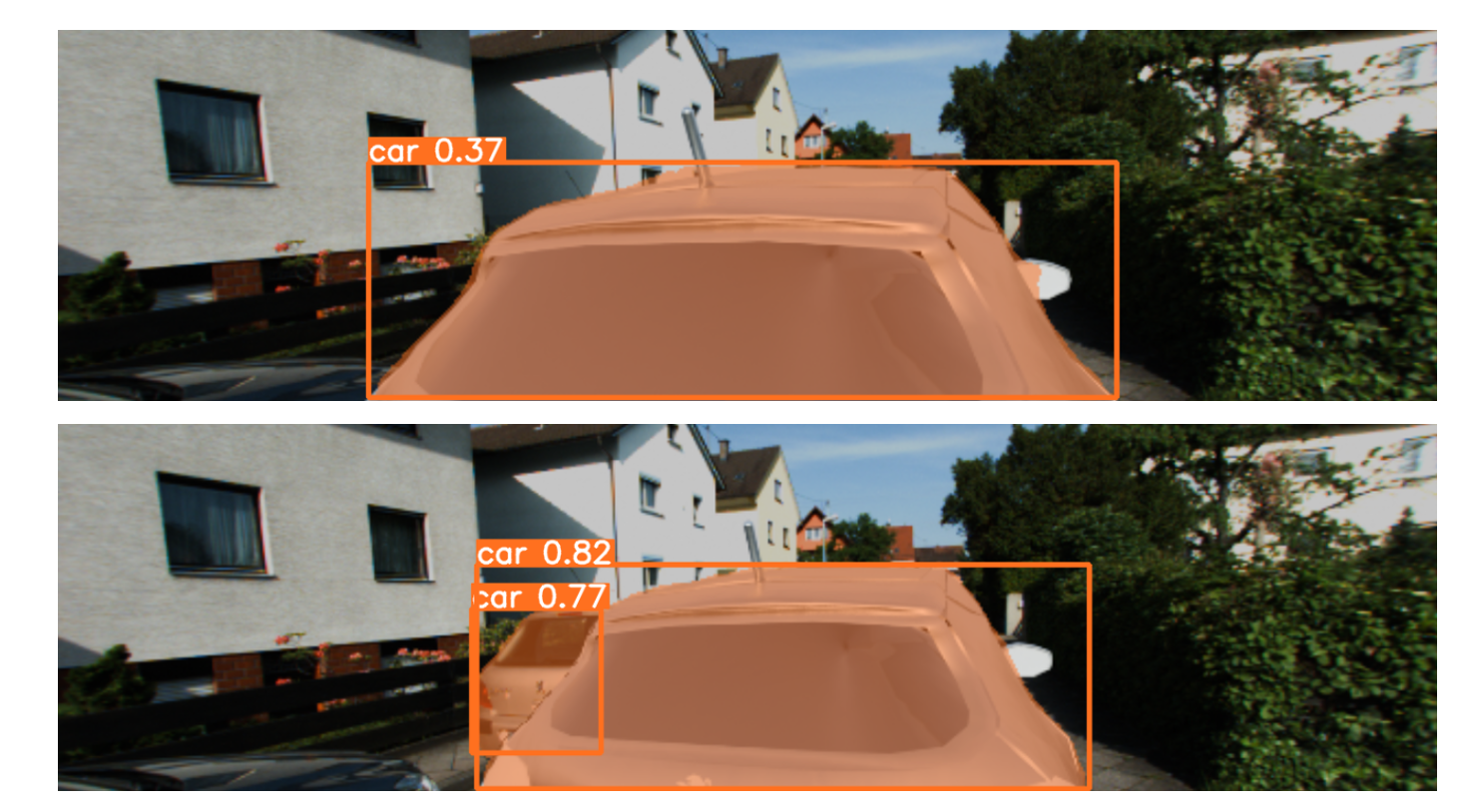}  
        \caption{Scenarios with CAD model with opitmization}    
    \end{subfigure} 
    \caption{Quality results on self-driving perception system.}
    \label{fig:self-driving perception}
\end{figure} 

\begin{figure}[htbp]
    \centering
    \includegraphics[width=\linewidth]{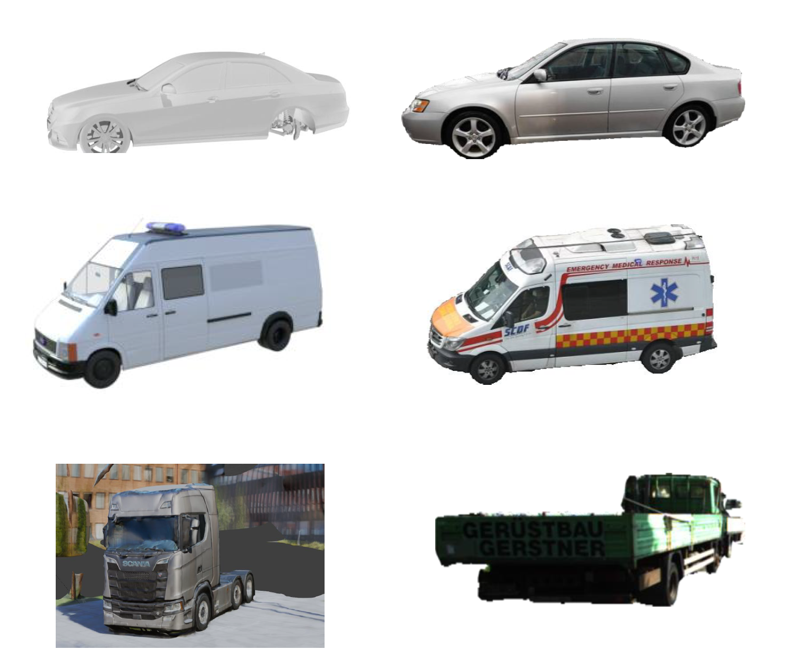}
    \caption{Failure Cases}
    \label{fig:failure_cases}
\end{figure}

\begin{figure*}[htbp]  
    \centering  
    \includegraphics[width=0.98\textwidth]{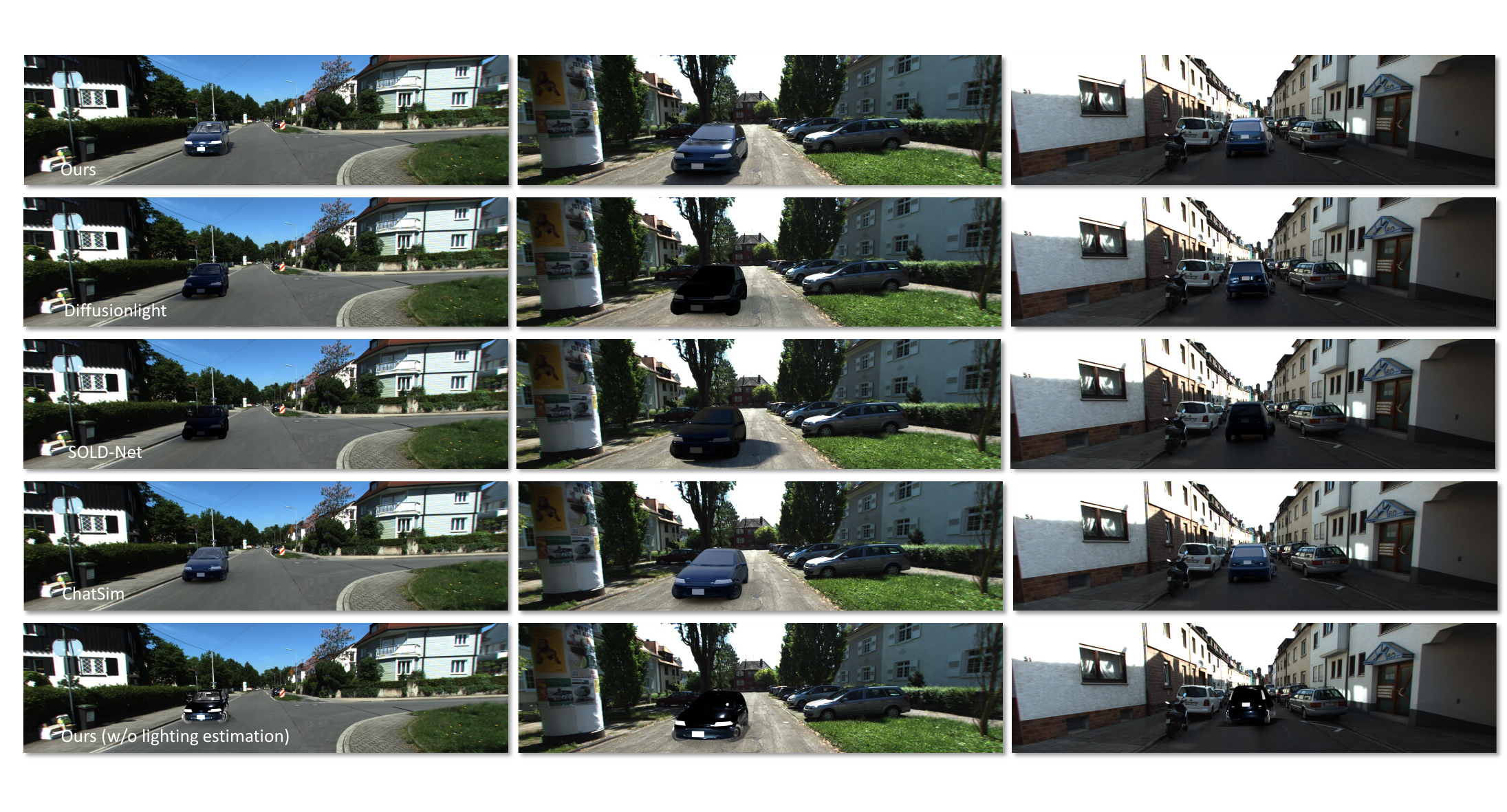} 
    \vspace{-1em}
    \caption{Lighting estimation comparison between ours, DiffusionLight~\cite{Phongthawee2023DiffusionLightLP}, ChatSim~\cite{Wei2024EditableSS}, SOLD-Net~\cite{Tang2022EstimatingSL}, and ours without lighting estimation. In the setup of ours (w/o lighting estimation), the vehicles are illuminated by six point lights positioned along the positive and negative x, y, and z axes. The results show that our method estimates environmental lighting more accurately, particularly in sunny weather.}  
    \label{fig:light_compare}  
\end{figure*} 

We conduct lighting estimation comparison experiments with three baselines as shown in \figref{fig:light_compare}. (1) lighting estimation method using the generative model: DiffusionLight~\cite{Phongthawee2023DiffusionLightLP}. (2) lighting estimation method with the auto-regressive network: SOLD-Net ~\cite{Tang2022EstimatingSL}. (3) lighting estimation method using ray-tracing: ChatSim ~\cite{Wei2024EditableSS}. For the DiffusionLight, we use the open-sourced official code and checkpoints and take the single-view perspective image as input. For the SOLD-Net, we manually select two points on the ground to mark the area where the network estimates the lighting. After obtaining the output results, we selected the HDR image that closely matched the lighting of the real scene for testing. For the ChatSim ~\cite{Wei2024EditableSS}, we used the view directly ahead of the vehicle as the network input. We also present the quality results of UrbanCAD without lighting estimation in \tabref{tab:self-driving}, where we use six point lights positioned in the positive and negative x, y, and z axis.
As demonstrated in the \figref{fig:light_compare}, our method performs better than the baselines, especially in sunny weather where the sun is absent from the perspective images. This is because our fisheye-based method has a 360$^\circ$ view of the environment to accurately capture the location and existence of the sun. However, our method may have limitations in estimating the lighting for objects in the shadow. This is due to the presence of overexposed areas in the fisheye camera's captured image. When these overexposed areas are combined into a panorama, they are given higher brightness, resulting in artifacts when lightening the vehicles in shadow in the final rendering.

\subsection{Quality results of perception system} \label{sec:perception}

In \figref{fig:self-driving perception}, we show the quality result of different perception results on synthetic data created by UrbanCAD (Ours) and UrbanCAD without material optimization. We find the perception system may fail to work in the synthetic data constructed with the vehicle models with unrealistic materials.

\begin{table}[t]
    \centering
    \begin{tabular}{c|cc}
    \toprule
    Method & FID$\downarrow$ & KID$\downarrow$ \\
    \midrule
    NeRS~\cite{Zhang2021NeRSNR} (Surrounding) & 110.55 & 0.0780 \\
    NeRS~\cite{Zhang2021NeRSNR} (Partial) & 206.46 & 0.1685 \\
    UrbanCAD (Ours) & 79.50 & 0.0530 \\
    \bottomrule
    \end{tabular}
    \vspace{-2mm}
    \captionof{table}{\textbf{Quantitative comparison} to NeRS on MVMC dataset in both surrounding and partial observation. Note that our method uses only a single-view image as input.}
    \label{tab:ners}
\end{table}

\subsection{Failure Cases} \label{sec:failure_case}

we provide failure cases in \figref{fig:failure_cases}. Our method may provide unsatisfactory results when the retrieved CAD model is defective (e.g., missing wheels), when the reference vehicle has multiple colors in one component (e.g., ambulance), or when the vehicles in the reference view are rarely seen (e.g., heavy-duty truck).

\begin{figure}[h]
    \centering
    \begin{tabular}{c|ccc}
    \toprule
     & Chamfer Dist.$\downarrow$ & Volume IOU$\uparrow$\\
     \midrule
    Ours &  \textbf{0.052} & \textbf{0.636}\\
    \midrule
    Wonder3D & 0.058 & 0.588\\
    \bottomrule
    \end{tabular}
    \captionof{table}{Geometry quality}
    \label{tab:geo_quality}
\end{figure}

\subsection{Geometry quality.} We randomly select 30 vehicles from the ShapeNet dataset, encompassing various types, and retrieve their corresponding models from the Objaverse dataset. We report the Chamfer Distance and Volume IOU in \tabref{tab:geo_quality}. Our retrieved models' geometry quality surpasses the reconstruction baseline, Wonder3D [42], as our retrieved CAD models often exhibit better geometry quality in unobservable regions.

\section{Broader Impact} \label{sec:social}

UrbanCAD may help the development of self-driving simulation technology, which can further encourage the development of the self-driving industry. However, our method may be used to create some false urban scenes, leading to some social problems.

\end{document}